\pdfoutput=1
\documentclass[10pt,twocolumn,letterpaper]{article}
\usepackage[pagenumbers]{cvpr}   

\usepackage{amsthm}
\usepackage{mathtools}
\usepackage{multirow}
\usepackage{tabularx}
\usepackage{makecell}
\usepackage{pifont}
\usepackage{tikz}
\usetikzlibrary{arrows.meta,positioning}

\newtheorem{proposition}{Proposition}
\newtheorem{lemma}{Lemma}
\theoremstyle{remark}

\usepackage{microtype}
\newcommand{\compactequations}{%
  \setlength{\abovedisplayskip}{2pt}%
  \setlength{\belowdisplayskip}{2pt}%
  \setlength{\abovedisplayshortskip}{2pt}%
  \setlength{\belowdisplayshortskip}{2pt}%
}

\AddToHook{cmd/normalsize/after}{\compactequations}
\AtBeginDocument{\compactequations}

\DeclareMathOperator{\LL}{LL}
\DeclareMathOperator{\Var}{Var}

\newcommand{\method}{\textup{\textsc{ContLoRA}}}
\newcommand{\overviewref}[1]{\cref{fig:overview}\textup{(#1)}}
\newcommand{\descriptorref}[1]{\cref{fig:descriptor-geometry}\textup{(#1)}}
\newcommand{\gmm}{\method\text{-GMM}}
\newcommand{\gmmf}{\method\text{-GMM+F}}
\newcommand{\gmmjl}{\method\text{-GMM+JL}}
\newcommand{\gmmjlf}{\method\text{-GMM+JL+F}}
\newcommand{\kcjl}{\method\text{-kC+JL}}
\newcommand{\chjl}{\method\text{-CH+JL}}
\newcommand{\hil}{\method\text{-Hil}}
\newcommand{\hilf}{\method\text{-Hil+F}}
\newcommand{\hiljlf}{\method\text{-Hil+JL+F}}
\newcommand{\cmark}{\ding{51}}
\newcommand{\xmark}{\ding{55}}
\newcommand{\takeaway}[1]{\noindent\textbf{Takeaway.} \emph{#1}}

\newif\ifdraft
\drafttrue
\ifdraft

\else

\fi

\newcommand{\novresult}[3]{\csname nov:#1:#2:#3\endcsname}
\expandafter\def\csname nov:cifar100:maxLL:auroc\endcsname{75.6{\pm}4.4}
\expandafter\def\csname nov:cifar100:maxLL:fpr95\endcsname{73.8{\pm}6.5}
\expandafter\def\csname nov:cifar100:mahalanobis:auroc\endcsname{72.2{\pm}6.0}
\expandafter\def\csname nov:cifar100:mahalanobis:fpr95\endcsname{76.6{\pm}8.3}
\expandafter\def\csname nov:cifar100:rmd:auroc\endcsname{88.7{\pm}2.7}
\expandafter\def\csname nov:cifar100:rmd:fpr95\endcsname{55.2{\pm}8.8}
\expandafter\def\csname nov:cifar100:energy:auroc\endcsname{89.5{\pm}1.4}
\expandafter\def\csname nov:cifar100:energy:fpr95\endcsname{53.1{\pm}2.9}
\expandafter\def\csname nov:cifar100:knn:auroc\endcsname{71.8{\pm}4.6}
\expandafter\def\csname nov:cifar100:knn:fpr95\endcsname{86.1{\pm}3.6}
\expandafter\def\csname nov:cifar100:lambda:auroc\endcsname{85.7{\pm}2.6}
\expandafter\def\csname nov:cifar100:lambda:fpr95\endcsname{66.5{\pm}2.2}
\expandafter\def\csname nov:imagenet_r:maxLL:auroc\endcsname{60.5{\pm}1.4}
\expandafter\def\csname nov:imagenet_r:maxLL:fpr95\endcsname{90.9{\pm}0.8}
\expandafter\def\csname nov:imagenet_r:mahalanobis:auroc\endcsname{59.6{\pm}1.3}
\expandafter\def\csname nov:imagenet_r:mahalanobis:fpr95\endcsname{89.3{\pm}1.0}
\expandafter\def\csname nov:imagenet_r:rmd:auroc\endcsname{76.8{\pm}1.0}
\expandafter\def\csname nov:imagenet_r:rmd:fpr95\endcsname{79.8{\pm}2.6}
\expandafter\def\csname nov:imagenet_r:energy:auroc\endcsname{74.3{\pm}1.0}
\expandafter\def\csname nov:imagenet_r:energy:fpr95\endcsname{88.4{\pm}2.5}
\expandafter\def\csname nov:imagenet_r:knn:auroc\endcsname{53.5{\pm}1.7}
\expandafter\def\csname nov:imagenet_r:knn:fpr95\endcsname{93.0{\pm}1.4}
\expandafter\def\csname nov:imagenet_r:lambda:auroc\endcsname{66.0{\pm}0.6}
\expandafter\def\csname nov:imagenet_r:lambda:fpr95\endcsname{89.8{\pm}1.2}

\usepackage{algorithm}
\usepackage{algorithmic}

\definecolor{cvprblue}{rgb}{0.21,0.49,0.74}
\usepackage[pagebackref,breaklinks,colorlinks,allcolors=cvprblue]{hyperref}
\usepackage[capitalize]{cleveref}

\def\confName{CVPR}
\def\confYear{2027}

\DeclareMathOperator{\Cov}{Cov}

\makeatletter
\renewcommand*\l@section{\@dottedtocline{1}{0em}{2.2em}}
\renewcommand*\l@subsection{\@dottedtocline{2}{1.5em}{3.2em}}
\makeatother

\title{The Geometry Is the Memory:\\Task-Density Routing for Continual LoRA}

\author{
Alireza Heidari\\
Elastic\\
{\tt\small alireza.heidari@elastic.co}
\and
Thomas Veasey\\
Elastic\\
{\tt\small tom.veasey@elastic.co}
\and
Mariano Tepper\\
Elastic\\
{\tt\small mariano.tepper@elastic.co}
}

\begin{document}
\maketitle
\addtocontents{toc}{\protect\setcounter{tocdepth}{-1}}


\begin{abstract}
Freezing older adapters does not ensure that a continual learner will select and use them properly at inference time. When multiple adapters exist, routing among them and identifying the adapter whose representation space matches the query’s answer can be difficult.
We propose \method, which replaces stored examples with compact class statistics for adapter selection and classification.
Gaussian class models in a frozen representation select LoRA adapters, while accumulated feature moments exactly recover a class-balanced ridge classifier for fixed training features.
Trained without retaining images or feature vectors, \method\ reaches $88.55\%$ accuracy on 20-task CIFAR-100 and $86.99\%$ on CUB-200.
Across three benchmarks in the matched component study, its mean final accuracy differs from an independently trained stored-feature counterpart by at most $0.07$ percentage points.
Ablations show that improved task coverage need not improve classification and adapter composition also matters.
On held-out classes, competitor-centered likelihood scores improve novelty AUROC over raw likelihood without altering routing ranks.
These results support statistical memory as an alternative to example-retention approaches in the evaluated settings, while identifying task ambiguity as an ongoing limitation.
\end{abstract}

\section{Introduction}
\label{sec:intro}

Class-incremental learning (CIL) requires a model to learn new
classes while retaining earlier capabilities, without task labels
at inference.
Pre-trained vision transformers provide transferable representations, and parameter-efficient methods adapt them through prompts, adapters, or low-rank updates~\cite{wang2022l2p,smith2023coda,
zhou2024ease,liang2024inflora,wu2025sdlora,zhu2025bilora,wang2025sema}.
However, preserving task-specific parameters does not guarantee
that the complete system continues to use them correctly.

Freezing a completed adapter prevents its weights from changing, but later tasks can still alter the global classifier and the adapters selected or combined at inference.
Retaining past features can re-balance the classifier, although
storing all such features increases memory as the stream grows.
This motivates two questions: 
\begin{itemize}
    \item \textit{\textbf{can compact statistics replace retained examples throughout a continual adapter system}}, and 
    \item \textbf{\textit{how do adapter selection, composition, and classifier alignment affect its performance?}}
\end{itemize}

\begin{figure*}[t]
\centering
\input{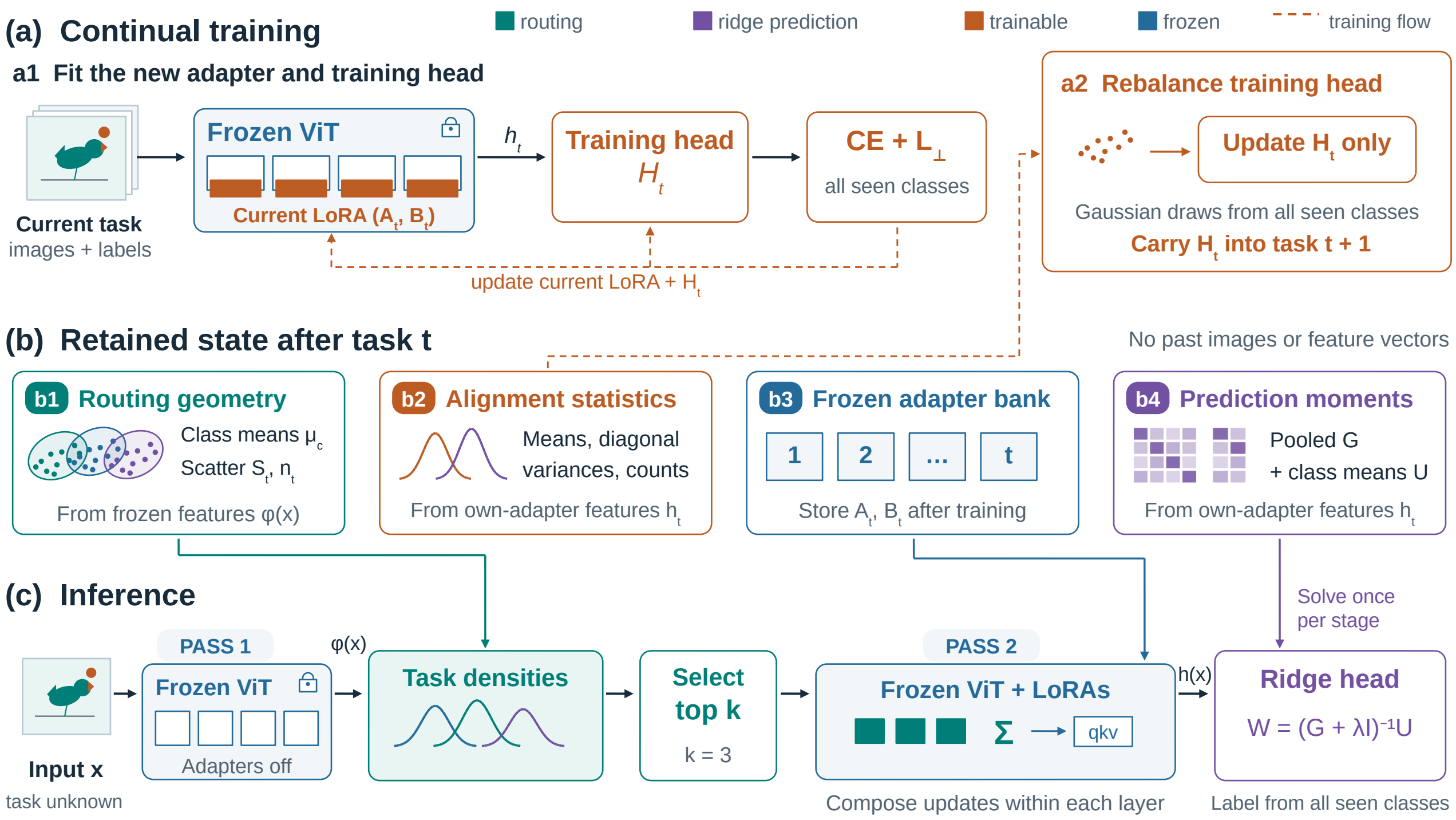}
\vspace{-2em}
\caption{Default \method\ pipeline.
\textbf{(a) Continual training.} Current-task data train the new adapter and global head (a1); Gaussian samples re-balance only the head, carried into the next task (a2).
\textbf{(b) Retained state.} After each stage, memory stores routing geometry (b1), alignment statistics (b2), frozen adapters (b3), and prediction moments (b4). Routing statistics use frozen features; alignment and prediction statistics use own-adapter features. Per-example data are discarded.
\textbf{(c) Inference.} A frozen pass selects adapters; a second pass composes their updates and predicts with a separate ridge head solved after each stage. Blocks are schematic; density colors distinguish distributions.}
\label{fig:overview}
\vspace{-1.5em}
\end{figure*}

\noindent We introduce \method, which uses class statistics as memory for
both adapter selection and classification (\cref{fig:overview}). Class means in a frozen representation, together with a covariance shared across tasks, define task densities that select the top-$k$ LoRA adapters.
Adapter training uses a linear head optimized with cross-entropy. After each stage, we re-balance this training head using Gaussian samples from stored class statistics and carry it into the next stage. For prediction, we solve a separate ridge head after each stage using accumulated moments of adapted features from all classes seen so far. These moments exactly recover the class-balanced ridge solution for the fixed training features; the system retains neither past images nor per-example feature vectors.

Our matched study on CIFAR-100, ImageNet-R, and CUB-200
(\cref{tab:main}) compares independently trained systems with
the same architecture, task orders, and evaluation configuration.
Training-head alignment uses either stored features or Gaussian samples from class statistics.
Across these benchmarks, replacing stored features with Gaussian samples for training-head alignment changes the final mean accuracy by at most $0.07$ percentage points. This supports statistical memory as an alternative to the retained
examples in the evaluated settings. Our ablations also show that better task coverage need not improve classification: adapter composition affects the result as well.

\method\ coordinates three uses of statistical memory: selecting adapters,
aligning the carried training head, and solving a separate prediction head.
The contribution is this architecture and a controlled test of replacing
retained features across its training and inference pipeline.

\noindent\textbf{Contributions and scope.}
\begin{itemize}[leftmargin=*,topsep=2pt,itemsep=1pt]

\item \textbf{Statistical memory for routing and classification.}
We design a system that retains class statistics instead of examples and compare it with an independently trained counterpart that uses stored features. 

\item \textbf{Analysis of density-based decisions.}
We relate the task-identification error to density overlap and model fit through a stage-wise bound, showing that it reflects genuine convergence rather than random behavior.
A variance identity characterizes when subtracting the mean
competitor log-likelihood reduces the score variance for a fixed task.
We also have separate measurements for downstream classification or novelty-detection performance guarantees.

\item \textbf{Controlled analysis of system components.}
We compare task descriptors, adapter selection, composition weights,
and classifier designs while holding adapter weights fixed.
Separate training runs assess the consequences of replacing
retained features during classifier alignment.

\end{itemize}
\section{Related Work}
\label{sec:related}

\noindent\textbf{Continual learning with pre-trained models.}
Parameter-efficient methods differ in how they select, share,
and combine adaptations to a pre-trained backbone.
L2P and DualPrompt use key-based prompt selection
\cite{wang2022l2p,wang2022dualprompt}, while CODA-Prompt combines
learned components through input-conditioned
weighting~\cite{smith2023coda}.
HiDe-Prompt explicitly separates task-identity inference from
within-task prediction~\cite{wang2023hide}.
Other approaches study compact parameter tuning and combinations
of feature subspaces~\cite{gao2023lae,zhou2024aper,zhou2024ease}.
LoRA-based methods constrain updates to reduce interference
\cite{liang2024inflora,zhang2025clora,wang2023olora}.
SD-LoRA decouples update magnitude and direction
\cite{wu2025sdlora}; BiLoRA constructs almost-orthogonal task
subspaces using two fixed bases~\cite{zhu2025bilora}.
SLAO studies continual merging into a single LoRA for language
models~\cite{qiao2025slao}, while mixture-of-adapter approaches
learn input-dependent composition
\cite{yu2024moeadapters,wang2025sema}.
\method\ addresses the inference memory surrounding frozen adapters:
it couples density selection with prediction statistics to distinguish
parameter preservation from successful use.

\noindent\textbf{Feature statistics as memory.}
Several exemplar-free methods replace stored samples with
summaries of their representations.
PASS augments prototypes with Gaussian noise~\cite{zhu2021pass},
FeCAM uses class covariance information for Mahalanobis
classification~\cite{goswami2023fecam}, and SLCA aligns a classifier
using Gaussian feature samples after slow backbone
adaptation~\cite{zhang2023slca}.
RanPAC combines random nonlinear projections with a classifier
computed from accumulated feature statistics
\cite{mcdonnell2023ranpac}.
\method\ separates the classifier carried through training from the head
used for prediction: Gaussian samples align the former, while exact moments
of own-adapter features determine the latter. Unlike RanPAC's projected
features, these prediction moments use adapted features directly, with
equal class weighting.

Representation summaries also support adapter selection.
SEMA uses representation descriptors for expansion and a learned
router for adapter composition~\cite{wang2025sema}.
HESTIA combines order-invariant linearized adaptation with
density-guided routing in task-free streams and analyzes retrieval
error through density quality and embedding
separability~\cite{le2026taskfree}.
\method\ uses class Gaussians with shared covariance in a frozen embedding
space~\cite{lee2018mahalanobis,goswami2023fecam}, followed by top-$k$
composition within attention-layer LoRA. It assumes known training boundaries
and learns no composition router. Our matched memory study tests selection, alignment, and prediction together;
fixed-adapter ablations expose when improved retrieval fails to improve
the composed classifier.

\noindent\textbf{Rehearsal.}
Experience replay~\cite{chaudhry2019tiny},
iCaRL~\cite{rebuffi2017icarl}, and DER++~\cite{buzzega2020der}
retain past examples.
We examine the memory needed for classifier alignment within the same
adapter architecture. \Cref{sec:geometry-vs-data} compares full and bounded
feature replay with statistical alternatives; independent streams
test replacement throughout training. Replay-free here permits
statistic-generated features but retains no past images or per-example
vectors. This tests feature-memory substitution, not superiority over
raw-image replay methods.

\noindent\textbf{Likelihood scores and novelty detection.}
Deep generative models can assign higher likelihood to
out-of-distribution inputs than to inputs from their training
distribution~\cite{nalisnick2019deep}.
Likelihood ratios against a background model can reduce the
influence of confounding input statistics~\cite{ren2019likelihood}.
Relative Mahalanobis distance similarly adjusts class-conditional
scores using a class-agnostic Gaussian background
\cite{ren2021rmd}.
Our adaptation of this correction uses the other stored task densities
as the background (\cref{sec:calibration}): subtracting their mean
log-likelihood forms a log ratio to their unnormalized geometric mean.
Unlike a separately fitted background, this construction reuses the router's
existing models. We distinguish its cross-input novelty-scoring effect
from adapter selection: centering preserves every within-input task ranking.

\section{Method}
\label{sec:method}
\Cref{fig:overview} presents the \method~architecture and functionalities, categorized into three main regimes: (a) continual training (a1--a2 training steps), (b) retained state (b1--b4 memory modules), and (c) inference. 

\subsection{Setting}
Tasks $t=1,\dots,T$ have known training boundaries and disjoint
label sets $\mathcal{Y}_t$. At stage $t$, evaluation covers all seen classes $\mathcal{Y}_{1:t}=\bigcup_{s\le t}\mathcal{Y}_s$ without task identity. The frozen ViT-B/16~\cite{dosovitskiy2021vit} has 12 transformer blocks and produces an adapter-free \texttt{[CLS]} embedding $\phi(\mathbf{x})\in\mathbb{R}^{d}$, $d=768$.
The default system retains no past images or per-example features;
replay-free here permits synthetic features drawn from statistics.
Dataset splits and task counts appear in \cref{sec:setup}.

The default \method\ combines Gaussian-sample training-head alignment,
shared-covariance routing, and ridge prediction.
The \emph{stored-feature training} variant instead aligns its training
head on retained adapted features, keeping the same inference design.
The \emph{GMM + linear head} variant uses those stored-feature-trained
adapters with per-task diagonal mixtures and the re-balanced linear
head for prediction. The latter two configurations reuse the same saved adapter weights, but differ in their router and prediction head.

\subsection{Isolated attention-layer adapters}
\label{sec:adapters}
Each task adds $\Delta W_t^l=B_t^lA_t^l/r$~\cite{hu2022lora}
to the \texttt{qkv} projection of all 12 transformer blocks, with
$A_t^l\in\mathbb{R}^{r\times d}$, $B_t^l\in\mathbb{R}^{3d\times r}$,
and $r=8$. We initialize $A_t^l$ with Gaussian entries and $B_t^l=0$;
the backbone and completed adapters remain frozen.

In \overviewref{a1}, current-task data train the new adapter and a
global linear head using $\mathcal{L}_t=\mathcal{L}_{\mathrm{CE},t}
+\mathcal{L}_{\perp}$. Cross-entropy uses all seen classes, with only
the current adapter active at full weight. The penalty discourages
overlap between new and completed adapters' output directions:
\begin{equation}
\label{eq:ortho}
\mathcal{L}_{\perp}
=\lambda_{\perp}\sum_l\sum_{s<t}
\big\|(B_t^l)^{\top}B_s^l\big\|_F^2,
\qquad\lambda_{\perp}=0.1.
\end{equation}
Only current $B_t^l$ factors receive penalty gradients; cross-entropy
updates current $A_t^l$, $B_t^l$, and the head.

The penalty is zero at the first task because the sum is empty.
It is used only during adapter training, not head re-balancing or inference.
It is soft and norm-dependent:
for bounded input-factor norms it controls an upper bound on update
overlap, without guaranteeing interference-free predictions
(Supp.~\cref{sec:forgetting}). No forgetting-rate guarantee follows.
Completed adapters enter the frozen
bank in \overviewref{b3}.

\subsection{Task densities and routing}
\label{sec:routing}
\noindent\textbf{Stored statistics and task densities.}
In \overviewref{b1}, we store each class's frozen-feature mean
$\boldsymbol{\mu}_c$ and accumulate within-class scatter $S_t$ and
sample count $n_t$. Each task is an equal-weight mixture of its class
Gaussians with a covariance shared across all seen classes:
\begin{equation}
\label{eq:density}
\begin{aligned}
\hat p_{j,t}(\mathbf{z})
&=\frac{1}{|\mathcal{Y}_j|}\sum_{c\in\mathcal{Y}_j}
\mathcal{N}\big(\mathbf{z}\mid\boldsymbol{\mu}_c,\hat\Sigma_t\big),\\
\hat\Sigma_t
&=(1-\gamma)\frac{S_t}{n_t}
+\gamma\frac{\operatorname{tr}(S_t/n_t)}{d}I.
\end{aligned}
\end{equation}
Here $\hat\Sigma_t$ estimates the pooled within-class covariance.
Shrinkage toward a scaled identity stabilizes its inversion;
$\gamma$ controls the amount, and we set $\gamma=0.1$.
The quadratic term in $\mathbf{z}$ is common to all tasks, so their
log-densities reduce to log-sum-exp evaluations of affine class scores,
minus $\log|\mathcal{Y}_j|$, up to a common input-dependent term.
For our equal-sized task label sets, ranking omits both common terms.
Unequal sets require retaining $-\log|\mathcal{Y}_j|$
(Supp.~\cref{sec:router-evaluation}).
Updating the covariance can change old likelihoods; new competitors
can change selection even with a fixed feature map.

\noindent\textbf{Alternative descriptor.}
A per-task diagonal GMM models the frozen embedding distribution
without class assignments, with a separate diagonal covariance per
component. This is the router of the \emph{GMM + linear head}
configuration. We fit $K=100$ components by EM~\cite{dempster1977em},
using $k$-means++~\cite{arthur2007kmeanspp}, at most 30 iterations,
and variance floor $10^{-6}$ ($615$\,KB/task).
\Cref{sec:ablation} compares the two descriptors' top-1 routing accuracy,
top-3 task coverage, and downstream classification accuracy.

\noindent\textbf{Adapter selection and composition.}
Define $\LL_j(\mathbf{x})=\log\hat p_{j,t}(\phi(\mathbf{x}))$ and
let $\mathcal{S}_k(\mathbf{x})$ contain the $\min(k,t)$ highest-scoring
tasks. Their updates are composed inside each adapted layer:
\begin{equation}
\label{eq:forward}
W^l(\mathbf{x})=W_0^l+
\frac{w_t}{r}\sum_{j\in\mathcal{S}_k(\mathbf{x})}B_j^lA_j^l.
\end{equation}
Inference uses a frozen pass to select adapters and one adapted pass
to compose them (\overviewref{c}, Passes~1--2), regardless of $k$.
We set $k=3$, with $w_t=1/3$ on CIFAR-100 and
$w_t=1/\min(k,t)$ on ImageNet-R and CUB-200. These coincide from
stage three onward; CIFAR-100 uses less total adapter weight at the
first two stages.

\noindent\textbf{Margin-gated ablation.}
Only at the final stage, the \emph{GMM + linear head} ablation
switches between one full-weight adapter and top-3 composition
using the top-two likelihood margin. Its threshold is calibrated on
synthetic mixture samples, without test data
(Supp.~\cref{sec:gate}). The default does not use this gate.

\subsection{Classifier: training head and ridge head}
\label{sec:alignment}
\noindent\textbf{Training head.}
Current-task training biases the global linear head toward new classes.
To restore access to old classes, the alignment memory (\overviewref{b2}) stores each class's
adapted-feature mean $\mathbf{m}_c$, diagonal variance $\mathbf{v}_c$,
and count $N_c$, using its own adapter at full weight.
After each stage, we draw $N_c$ samples from
$\mathcal{N}(\mathbf{m}_c,\operatorname{diag}(\mathbf{v}_c))$
for every seen class and re-balance only the head
(\overviewref{a2}), following Gaussian alignment~\cite{zhang2023slca}.
The draw is reused for 30 SGD epochs; settings are in
Supp.~\cref{sec:hyperparameters}.
Sampling preserves original class counts rather than imposing equal
class frequencies. The head carries into the next stage to train the
next adapter; real features and synthetic samples are discarded.
Own-adapter feature maps stay fixed because their backbones and
completed adapters are frozen.

\noindent\textbf{Ridge prediction head.}
Prediction uses a separate class-balanced least-squares head.
Let $\tilde{\mathbf{h}}=[\mathbf{h};1]$ augment the own-adapter
training feature with a bias coordinate, and let
$\widehat{\mathbb E}_c$ average over class $c$'s training examples.
The prediction memory (\overviewref{b4}) retains
\[
G=\sum_{c\in\mathcal{Y}_{1:t}}
\widehat{\mathbb E}_c[\tilde{\mathbf{h}}\tilde{\mathbf{h}}^\top],
\qquad U=[\widehat{\mathbb E}_c[\tilde{\mathbf{h}}]]_{c\in\mathcal{Y}_{1:t}}.
\]
For $\lambda>0$, $W=(G+\lambda I)^{-1}U$ uniquely minimizes
\[
\sum_{c\in\mathcal{Y}_{1:t}}
\widehat{\mathbb E}_c\!\left[
\|W^\top\tilde{\mathbf{h}}-\mathbf{e}_c\|_2^2\right]
+\lambda\|W\|_F^2,
\]
where $\mathbf{e}_c$ is a one-hot target. Classes contribute equally,
and the regularizer includes the bias row. Prediction takes
$\arg\max_c(W^\top\tilde{\mathbf{h}}(\mathbf{x}))_c$.
These moments exactly recover the retained-feature ridge solution
for the same fixed features, weights, and regularization; no Gaussian
assumption is needed (Supp.~\cref{sec:ridge-proof}).
This does not remove the mismatch between own-adapter training
features and composed test features or guarantee accuracy.
We set $\lambda=0.01\operatorname{tr}(G)/(d+1)$ using held-out
training features from one order of dataset.
This analytic approach follows RanPAC~\cite{mcdonnell2023ranpac}
without its random projection.

\subsection{Competitor-centered novelty scores}
\label{sec:calibration}
Routing compares tasks within an input; novelty scoring compares
inputs. Input-dependent variation shared across task likelihoods can
obscure novelty~\cite{nalisnick2019deep,ren2019likelihood}.
A common constant offset cannot change rankings across inputs.
For $t>1$, define
\begin{equation}
\label{eq:lr}
\Lambda_j(\mathbf{x})=\LL_j(\mathbf{x})
-\frac{1}{t-1}\sum_{q\ne j}\LL_q(\mathbf{x}).
\end{equation}
Equivalently,
$\Lambda_j=\frac{t}{t-1}\LL_j-\frac{1}{t-1}\sum_q\LL_q$.
Thus centering preserves every within-input task ranking and top-$k$
set, while potentially changing rankings across inputs.
Our offline novelty study scores $\max_j\Lambda_j$ using per-task
GMMs, with larger scores indicating known classes, and reports
AUROC/FPR95. It implements neither deployment threshold calibration
nor a conformal detector.
Centering also rescales the top-two margin and its matched quantile
threshold by $t/(t-1)$, leaving the gated ablation unchanged.
The supplement states the separate conformal assumptions.

\subsection{What is stored}
\label{sec:memory}
\Cref{tab:memory} gives packed fp32 representation sizes in decimal
KB/MB, excluding the common backbone, optimizer, and temporary state.
On 20-task CIFAR-100, adapters occupy $23.59$\,MB; router and ridge
moments each occupy $1.49$\,MB in distinct feature spaces; alignment
statistics occupy $0.61$\,MB and the training head $0.31$\,MB.
Caching derived ridge weights adds $0.31$\,MB.
These are storage estimates: the implementation uses full fp64
matrices and solver workspace, detailed in the supplement.

\begin{table}[t]
\centering
\caption{
Retained-state accounting for ViT-B/16 with $r=8$ on
CIFAR-100 with five classes per task.
Sizes use fp32, with symmetric matrices packed; counts and identifiers are separate.
Rows above the line describe default state. Below, the
\emph{GMM + linear head} configuration uses the GMM; both feature-replay
variants retain 500 features/class; the buffer-head ablation
(\cref{tab:alignment}) retains 20/class.
}

\label{tab:memory}
\footnotesize
\setlength{\tabcolsep}{2pt}
\begin{tabular}{lrc}
\toprule
\textbf{Item} & \textbf{Size} & \textbf{Retained form} \\
\midrule
LoRA adapter (12 blocks, \texttt{qkv})
    & 1.18\,MB/task & parameters \\
Router: class means
    & 15\,KB/task & means \\
Router: shared scatter (once)
    & 1.18\,MB & scatter \\
Training head: class statistics
    & 31\,KB/task & mean, variance \\
Training head: weights and bias
    & 15\,KB/task & parameters \\
Ridge head: class means $\mathbf{u}_c$
    & 15\,KB/task & moments \\
Ridge head: Gram matrix $G$ (once)
    & 1.18\,MB & moments \\
Counts and class/task identifiers
    & $O(C+T)$ & integers \\
\midrule
Per-task GMM ($K=100$)
    & 615\,KB/task & parameters \\
Stored adapted features
    & 7.7\,MB/task & features \\
Feature buffer, 20/class
    & 0.31\,MB/task & features \\
\bottomrule
\end{tabular}
\end{table}
\section{Analysis}
\label{sec:analysis}
We characterize routing error and score variance, separately from
downstream classification.

\noindent\textbf{Routing error from stored densities.}
Fix a stage $t$. The fitted densities $\hat p_j=\hat p_{j,t}$ induce
distributions $\hat P_j$; $P_i$ is the true distribution of frozen
embeddings $\phi(\mathbf{x})$ for task $i$.
Define their fitted overlap as
$\mathrm{OVL}(\hat p_i,\hat p_j)=\int_{\mathbb R^d}\min\{\hat p_i(\mathbf{z}),\hat p_j(\mathbf{z})\}\,d\mathbf{z}$.
Condition on the fitted mixtures and take the test input independent of the fitting data.
Let $E_i=\bigcup_{j\ne i}\{\hat p_j\ge\hat p_i\}$ and $\delta_i=|P_i(E_i)-\hat P_i(E_i)|$.

\begin{proposition}
\label{prop:routing}
For a task-$i$ input, top-1 likelihood routing over $t$ tasks errs with $P_i$-probability at most
$\sum_{j\neq i}\mathrm{OVL}(\hat p_i,\hat p_j) + \delta_i$.
\end{proposition}
\noindent A union bound under $\hat P_i$ and transfer by $\delta_i$
give the result for any tie-breaking rule (Supp.).

\noindent The event-specific $\delta_i$ avoids a vacuous TV bound:
deterministic embeddings of discrete 8-bit images have atomic $P_i$,
whereas $\hat P_i$ has a density, giving $\mathrm{TV}(P_i,\hat P_i)=1$.
Estimate its two event probabilities from held-out task-$i$ data and
independent $\hat P_i$ samples, respectively, then take their absolute
difference. Sampling-error control is required for a certificate;
none is reported (Supp.).
Updating the shared covariance changes old densities, overlaps, and
$E_i$; the bound applies anew at each stage, rather than merely adding
an overlap term. It does not bound classification error.

\noindent\textbf{Variance of competitor centering.}
For $t>1$, let $X$ follow a fixed reference population and condition
on the fitted models. Suppose $\{\LL_j(X)\}$ have common variance
$\sigma^2>0$ and common pairwise correlation $\rho$.
\begin{lemma}
\label{lem:var}
$\Var(\Lambda_j(X))=\sigma^2\,\frac{t(1-\rho)}{t-1}$.
\end{lemma}
\noindent Expanding the variance gives the result (Supp.). It is smaller than $\sigma^2$ only when $\rho>1/t$.
The identity concerns a fixed task index, not the selected maximum, and it gives no guarantee for mean separation, AUROC, or false-positive rate.
We therefore evaluate novelty scores directly instead of inferring detection quality from variance reduction.

\section{Experiments}
\label{sec:experiments}

\subsection{Setup}
\label{sec:setup}
\noindent\textbf{Benchmarks.}
For the component study, we split CIFAR-100~\cite{krizhevsky2009cifar} into 20 tasks of 5 classes and 10 tasks of 10 classes, ImageNet-R~\cite{hendrycks2021imagenetr} into 10 tasks of 20 classes and 25 tasks of 8 classes, with a fixed per-class 80/20 train/test split (approximately 24k / 6k images), and CUB-200-2011~\cite{wah2011cub} into 10 tasks of 20 classes with its official split (5{,}994 / 5{,}794 images).

\noindent\textbf{Model and training.}
ViT-B/16 pre-trained on ImageNet-21k (AugReg~\cite{steiner2022augreg}), frozen; rank-8 LoRA on \texttt{qkv} in all 12 transformer blocks; AdamW, learning rate $3{\times}10^{-4}$, cosine decay, batch 64, 15 epochs per task at $T{=}25$ and $T{=}20$, and 20 epochs per task at $T{=}10$.
\method\ trains with the sampled-statistics head, routes with~Eq.\eqref{eq:density} and $k{=}3$, predicts with the ridge head, and retains no images or features.
A second \method\ run per class order uses the stored-feature head instead and carries auxiliary heads for \cref{sec:geometry-vs-data}.

\noindent\textbf{Comparison systems.}
We compare two local summed-adapter controls with the same backbone and optimizer; both re-balance their head on stored features, i.e., they use feature replay.
\emph{Sum, fixed-$A$} fixes a random orthonormal $A_t$, trains $B_t$, freezes completed modules, and sums them at inference. \emph{Sum, trainable-$B$} freezes previously learned $A$ factors while continuing to update their $B$ factors. These methods capture the core ideas of BiLoRA and SD-LoRA; however, for an exact comparison, we use the official published implementations; \emph{Sum, fixed-$A$} omits BiLoRA’s two-basis construction, while \emph{Sum, trainable-$B$} omits SD-LoRA’s fixed-direction/learned-magnitude parameterization.
CIFAR summed controls combine four archived runs with a new seed-1993 run; other continual runs in \cref{tab:main} use the new runner.
An archived vision adaptation of SLAO~\cite{qiao2025slao} is not used as a comparator because its initialization departs from the published rule (Supp.~\cref{sec:baselines}).
Separate \method\ runs appear in \cref{tab:crossbench}. Its ImageNet-R rows use 25 tasks of 8 classes, separately from the 10-task of \cref{tab:main}. 

\noindent\textbf{Metrics.}
$A_{\mathrm{last}}=T^{-1}\sum_{j=1}^T R_{T,j}$: mean final per-task accuracy. This equals image accuracy on equal-sized CIFAR-100 tasks; ImageNet-R and CUB-200 tasks have unequal test counts, so there it is a task-macro average.
Forgetting $\mathrm{Fgt}=\frac{1}{T-1}\sum_{j<T}\big(\max_{j\le i\le T}R_{i,j}-R_{T,j}\big)$, where $R_{i,j}$ is accuracy on task $j$ after stage $i$~\cite{chaudhry2018rwalk}.
Task-0: accuracy on the first task after the last stage, $R_{T,1}$.
Routing accuracy: fraction of test inputs whose true task is the top-1 (or in the top-$k$) of $\{\LL_j\}$.
The component study uses five CIFAR-100 class-order seeds $\{0,7,42,1993,2024\}$ and three ImageNet-R and CUB-200 seeds $\{7,42,2024\}$; $\pm$ denotes a 95\% Student-$t$ interval.
Paired differences use shared class orders; this pairing does not remove implementation or tuning confounds. Tests are exploratory and uncorrected for multiple comparisons.

\subsection{Main results}
\label{sec:main}

\begin{table*}[t]
\centering
\caption{Class-incremental results with rank-8 LoRA in all attention blocks of a frozen ViT-B/16. \emph{Replay}: whether the head is re-balanced on stored adapted features. \method\ rows use the shared-covariance router and ridge head; the \emph{GMM + linear head} configuration uses per-task diagonal mixtures and a stored-feature linear head. The oracle selects the true task's adapter at full weight. Continual runs: 5 class orders on CIFAR-100, 3 on ImageNet-R and CUB-200; $\pm$: 95\% interval. Task-0 is the mean over orders. \textbf{Bold}: best continual mean and values within $0.1$\,pp, excluding joint and oracle rows; this is a display convention, not a significance test. Joint references train all classes together; see Supp. for their separate protocols.}

\label{tab:main}
\footnotesize
\setlength{\tabcolsep}{2.1pt}
\resizebox{\textwidth}{!}{%
\begin{tabular}{l c ccc ccc ccc}
\toprule
& & \multicolumn{3}{c}{\textbf{CIFAR-100, $T{=}20$}} & \multicolumn{3}{c}{\textbf{ImageNet-R, $T{=}10$}} & \multicolumn{3}{c}{\textbf{CUB-200, $T{=}10$}} \\
\cmidrule(lr){3-5}\cmidrule(lr){6-8}\cmidrule(lr){9-11}
\textbf{Method} & \textbf{Replay} & $A_{\mathrm{last}}\uparrow$ & Fgt$\downarrow$ & Task-0$\uparrow$ & $A_{\mathrm{last}}\uparrow$ & Fgt$\downarrow$ & Task-0$\uparrow$ & $A_{\mathrm{last}}\uparrow$ & Fgt$\downarrow$ & Task-0$\uparrow$ \\
\midrule
Joint (reference) & -- & $92.90{\pm}0.14$ & -- & -- & $81.69$ & -- & -- & $87.79{\pm}0.16$ & -- & -- \\
\midrule
Sum, trainable-$B$ & \cmark & $50.79{\pm}13.13$ & $50.80{\pm}13.92$ & $32.84$ & $49.63{\pm}1.64$ & $45.15{\pm}1.42$ & $32.02$ & $58.72{\pm}3.98$ & $41.58{\pm}3.31$ & $28.36$ \\
Sum, fixed-$A$ & \cmark & $84.55{\pm}1.36$ & $12.88{\pm}1.62$ & $55.48$ & $\mathbf{75.15{\pm}0.90}$ & $\mathbf{7.64{\pm}0.86}$ & $65.85$ & $86.30{\pm}0.88$ & $6.66{\pm}0.84$ & $75.99$ \\
\method, GMM + linear head & \cmark & $86.50{\pm}0.34$ & $6.93{\pm}0.62$ & $79.52$ & $70.52{\pm}0.22$ & $12.35{\pm}0.46$ & $65.17$ & $85.98{\pm}0.56$ & $5.41{\pm}0.62$ & $80.97$ \\
\method, stored-feature training & \cmark & $\mathbf{88.57{\pm}0.15}$ & $\mathbf{5.16{\pm}0.26}$ & $\mathbf{85.72}$ & $73.40{\pm}0.91$ & $8.51{\pm}1.80$ & $75.04$ & $\mathbf{87.05{\pm}0.15}$ & $\mathbf{4.68{\pm}0.56}$ & $83.49$ \\
\midrule
\method\ (replay-free) & \xmark & $\mathbf{88.55{\pm}0.19}$ & $\mathbf{5.16{\pm}0.36}$ & $\mathbf{85.68}$ & $73.33{\pm}0.94$ & $8.59{\pm}1.83$ & $\mathbf{75.09}$ & $\mathbf{86.99{\pm}0.26}$ & $\mathbf{4.68{\pm}0.83}$ & $\mathbf{83.61}$ \\
\quad\emph{oracle task identity} & \xmark & $96.90{\pm}0.21$ & $1.30{\pm}0.20$ & $96.12$ & $86.15{\pm}0.68$ & $2.51{\pm}0.25$ & $88.60$ & $89.81{\pm}0.19$ & $3.54{\pm}0.40$ & $87.84$ \\
\bottomrule
\end{tabular}}
\end{table*}

\begin{figure*}[t]
\centering
\includegraphics[width=\textwidth]{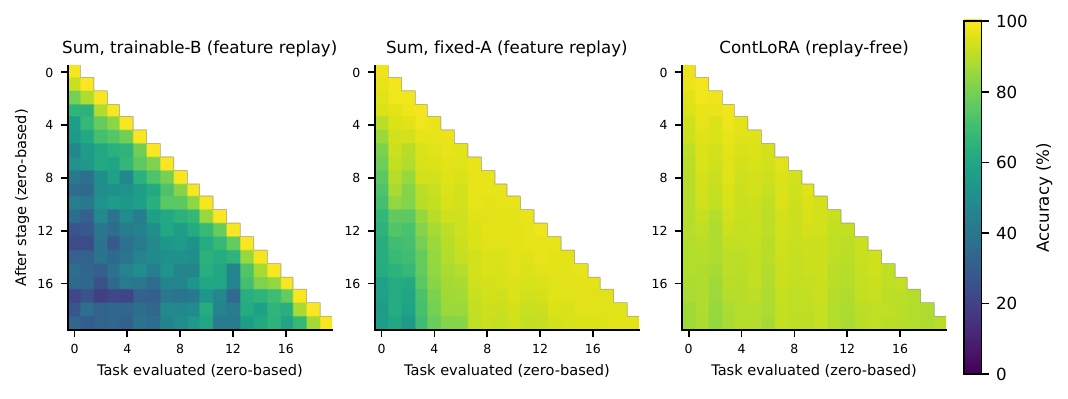}

\caption{Accuracy $R_{i,j}$ on task $j$ (columns) after stage $i$ (rows), CIFAR-100, $T{=}20$, mean over the five class orders of \cref{tab:main}, common color scale. The summed-adapter controls re-balance their head on stored features and still lose more early-task accuracy than the replay-free \method.}
\label{fig:heatmap}
\end{figure*}

\Cref{tab:main} reports the comparison systems.
On CIFAR-100 the replay-free \method\ reaches $88.55\%$, $4.01{\pm}1.31$\,pp above the fixed-$A$ control (paired over five orders, $p{=}0.001$), with forgetting $5.16\%$ versus $12.88\%$ and first-task accuracy $85.7\%$ versus $55.5\%$ (\cref{fig:heatmap}).
On CUB-200 it reaches $86.99\%$, $0.69$\,pp above the control on all three orders, reduces forgetting from $6.66\%$ to $4.68\%$, and raises first-task accuracy from $76.0\%$ to $83.6\%$.
The control re-balances its head on stored features, whereas \method\ retains none.
In the paired runs of \cref{tab:main}, training \method\ with stored features instead raises final accuracy by only $0.02$, $0.07$ and $0.06$\,pp on the three benchmarks.
Relative to the \emph{GMM + linear head} configuration, the replay-free system gains $2.05{\pm}0.44$, $2.80{\pm}0.98$ and $1.01{\pm}0.49$\,pp; \cref{sec:geometry-vs-data,sec:ablation} attribute most of this to the ridge head.
Joint training is a separate reference, not an upper bound; its capacity and optimization budget differ (Supp.). The CUB-200 joint mean is $0.80$\,pp above the default continual mean.

\noindent\textbf{ImageNet-R: selection and composition.}
Here \method\ reaches $73.33\%$, $1.82{\pm}1.73$\,pp below the fixed-$A$ control ($p{=}0.046$), with similar forgetting ($8.59\%$ versus $7.64\%$, $p{=}0.27$) and higher first-task accuracy ($75.1\%$ versus $65.9\%$).
Selecting the true task's adapter at full weight reaches $86.15\%$, $11$\,pp above the control, showing substantial sensitivity to adapter selection and composition.
The router ranks the true task first for $69.7\%$ of inputs and among the top three for $83.5\%$, compared with $88.0\%$/$96.2\%$ on CIFAR-100 and $89.7\%$/$98.1\%$ on CUB-200: renditions share styles across classes, and tasks are random class splits, so frozen features separate them less well.
Across benchmarks, the oracle-minus-default gap is $2.8$, $8.4$ and $12.8$\,pp on CUB-200, CIFAR-100 and ImageNet-R, in decreasing order of top-3 coverage.
The oracle supplies unavailable task identity and replaces top-3 composition by one full-weight adapter. Its gap combines selection and composition effects and does not isolate a recoverable routing loss.
Two checkpoint diagnostics with stored-feature-trained adapters did not help (Supp.): moment-based composition weights and scoring each shortlisted task's classes under its own adapter.

\begin{table}[t]
\centering
\caption{Additional runs of the cited methods on ViT-B/16, final accuracy (\%). ImageNet-R uses $T{=}25$ (8 classes per task), separately from $T{=}10$ in \cref{tab:main}; the \method\ rows are also separate runs. Baselines use method-specific class orders, backbone checkpoints and training budgets (Supp.). \textbf{Bold}: highest mean in each column.}
\vspace{-0.75em}
\label{tab:crossbench}
\footnotesize
\setlength{\tabcolsep}{3pt}
\resizebox{\columnwidth}{!}{%
\begin{tabular}{l ccc}
\toprule
& \textbf{CIFAR-100} & \textbf{ImageNet-R} & \textbf{CUB-200} \\
\textbf{Method} & $T{=}20$ & $T{=}25$ & $T{=}10$ \\
\midrule
L2P~\cite{wang2022l2p} & 79.51 & 63.26 & 67.13 \\
InfLoRA~\cite{liang2024inflora} & 80.97 & 68.65 & 70.47 \\
SD-LoRA~\cite{wu2025sdlora} & 84.14 & 68.34 & 77.48 \\
C-LoRA~\cite{zhang2025clora} & 86.70 & 70.14 & 82.25 \\
CL-LoRA~\cite{he2025cllora} & 85.32 & 70.73 & 80.03 \\
\midrule
\method\ (replay-free) & 88.52 & \textbf{71.92} & 86.86 \\
\method, stored-feature training & \textbf{88.69} & 71.87 & \textbf{87.12} \\
\bottomrule
\end{tabular}}
\vspace{-2em}
\end{table}

\noindent\textbf{Additional comparison runs.}
\Cref{tab:crossbench} compares \method\ with L2P, InfLoRA, SD-LoRA, C-LoRA and CL-LoRA on 20-task CIFAR-100, 25-task ImageNet-R and 10-task CUB-200.
These runs are distinct from the paired component study in \cref{tab:main}; their means should not be substituted into its uncertainty estimates or ablations.
The replay-free configuration exceeds the strongest listed baseline by $1.82$, $1.19$ and $4.61$\,pp, respectively.
Stored-feature training changes its accuracy by $+0.17$, $-0.05$ and $+0.26$\,pp compared to the replay free.
These are descriptive comparisons under method-specific training recipes; aggregate means alone do not establish statistical significance.
Supp.~\cref{sec:extended} compares geometric descriptors on frozen features under a separate known-vs-novel selection protocol, without attention-layer adapters or the ridge head.

\takeaway{Compact class statistics are a sufficient substitute for retained features: replacing all stored features changes final accuracy by less than a tenth of a point, and replay-free \method\ remains roughly $1$--$5$\,pp above the listed baselines. The remaining error arises from adapter selection and composition at inference rather than from a loss of task knowledge.}

\subsection{Geometry versus data}
\label{sec:geometry-vs-data}

\Cref{tab:alignment} uses the stored-feature training run, keeps its adapters and per-task mixture router fixed, and changes only the head; all rows share the same adapters.
The ridge head, built from class moments alone, is the most accurate: $+2.22{\pm}0.37$\,pp over the stored-feature head on CIFAR-100, $+2.20{\pm}0.69$ on ImageNet-R and $+1.08{\pm}0.56$ on CUB-200, with forgetting $5.03\%$ versus $6.93\%$ on CIFAR-100, while keeping $1.5$\,MB instead of $154$\,MB.
A linear head re-balanced on sampled statistics loses $2.64{\pm}0.58$\,pp against stored features on CIFAR-100, and without alignment the head reaches only $9.09\%$.
Solving ridge regression from exact moments requires neither Gaussian sampling nor iterative head re-balancing.
Whether the adapters themselves need stored features is answered by the independent replay-free run of \cref{tab:main}: with the ridge head it matches the stored-feature run within $0.07$\,pp on all three benchmarks, although its in-run sampled head is $5.3$\,pp lower on CIFAR-100.

\begin{table}[t]
\centering
\caption{$A_{\mathrm{last}}$ (\%) by what the head is built from (stored-feature training run; per-task mixture router, top-3; adapters and router shared across rows). Memory counts only alignment data for that head over all CIFAR-100 classes. The buffer stores features, not images; this is not an ER comparison.}

\label{tab:alignment}
\footnotesize
\setlength{\tabcolsep}{1.0pt}
\begin{tabular}{lrccc}
\toprule
\textbf{Head} & \textbf{Mem.} & \textbf{CIFAR-100} & \textbf{IN-R} & \textbf{CUB} \\
\midrule
Stored features & 154\,MB & $86.50{\pm}0.34$ & $70.52{\pm}0.22$ & $85.98{\pm}0.56$ \\
Buffer 20/class & 6.1\,MB & $82.36{\pm}0.58$ & $66.99{\pm}0.96$ & $85.47{\pm}0.93$ \\
\midrule
Ridge on moments & 1.5\,MB & $\mathbf{88.72{\pm}0.13}$ & $\mathbf{72.72{\pm}0.60}$ & $\mathbf{87.05{\pm}0.39}$ \\
Sampled statistics & 0.6\,MB & $83.86{\pm}0.55$ & $67.70{\pm}0.42$ & $85.84{\pm}1.02$ \\
Class means, NCM & 0.3\,MB & $81.76{\pm}0.27$ & $64.28{\pm}1.00$ & $85.86{\pm}0.25$ \\
None & 0 & $9.09{\pm}1.11$ & $12.81{\pm}0.95$ & $15.35{\pm}2.42$ \\
\bottomrule
\end{tabular}
\end{table}
\begin{table}[t]
\centering
\caption{Ablations, $A_{\mathrm{last}}$ (\%; 5 class orders on CIFAR-100, 3 on ImageNet-R and CUB-200). Top block: replay-free adapters with the ridge head. Middle block: the \emph{GMM + linear head} configuration (per-task mixtures, stored-feature head), shared adapters. Bottom: adapters retrained without the penalty of~Eq.\eqref{eq:ortho} (stored-feature head, 3 paired orders). Non-default middle-block rows re-evaluate the final model with fixed adapters and classifier.}
\vspace{-0.75em}
\label{tab:ablation}
\footnotesize
\setlength{\tabcolsep}{1.8pt}
\begin{tabular}{lccc}
\toprule
\textbf{Variant} & \textbf{CIFAR-100} & \textbf{IN-R} & \textbf{CUB-200} \\
\midrule
\multicolumn{4}{l}{\emph{Router (replay-free adapters, ridge head)}} \\
Oracle task identity & $96.90{\pm}0.21$ & $86.15{\pm}0.68$ & $89.81{\pm}0.19$ \\
Shared cov., top-1 & $88.38{\pm}0.12$ & $71.12{\pm}0.49$ & $88.10{\pm}0.02$ \\
Shared cov., top-3 (default) & $88.55{\pm}0.19$ & $73.33{\pm}0.94$ & $86.99{\pm}0.26$ \\
Per-task GMM $K{=}100$, top-3 & $88.71{\pm}0.15$ & $72.73{\pm}0.69$ & $86.99{\pm}0.49$ \\
\midrule
\multicolumn{4}{l}{\emph{Composition (GMM + linear head)}} \\
Top-1 & $85.88{\pm}0.20$ & $62.77{\pm}0.62$ & $86.31{\pm}0.28$ \\
Top-3, weights $1/3$ & $86.50{\pm}0.34$ & $70.52{\pm}0.22$ & $85.98{\pm}0.56$ \\
Top-3, weights $1$ & $85.01{\pm}0.87$ & $44.36{\pm}2.29$ & $83.72{\pm}1.17$ \\
Gated top-1/top-3 & $86.77{\pm}0.30$ & $70.44{\pm}0.50$ & $86.07{\pm}0.41$ \\
GMM $K{=}20$ instead of $100$ & $86.45{\pm}0.45$ & $70.38{\pm}0.85$ & $85.83{\pm}0.54$ \\
\midrule
\multicolumn{4}{l}{\emph{Adapters (retrained, seeds 7, 42, 2024)}} \\
With penalty, same seeds & $86.48{\pm}0.93$ & $70.44{\pm}1.02$ & $86.08{\pm}0.38$ \\
No $\mathcal{L}_\perp$ ($\lambda_\perp{=}0$) & $87.62{\pm}0.65$ & $69.11{\pm}0.71$ & $86.37{\pm}0.31$ \\
\bottomrule
\end{tabular}
\vspace{-1em}
\end{table}
\begin{table*}[t]
\centering
\caption{Known-vs-novel detection on frozen ViT-B/16 features (AUROC$\uparrow$ / FPR95$\downarrow$, mean $\pm$ 95\% interval over five class orders). Max-LL and $\Lambda$ use the per-task mixture densities; Mahalanobis uses class means and a tied full covariance; relative Mahalanobis additionally fits a global mean and covariance; energy uses a linear probe on the known classes; $k$-NN ($k{=}50$) keeps every training feature and is a reference only.}
\vspace{-0.75em}
\label{tab:novelty}
\small
\setlength{\tabcolsep}{8pt}
\begin{tabular}{lcccc}
\toprule
& \multicolumn{2}{c}{\textbf{CIFAR-100}} & \multicolumn{2}{c}{\textbf{ImageNet-R}} \\
\cmidrule(lr){2-3}\cmidrule(lr){4-5}
\textbf{Score} & AUROC$\uparrow$ & FPR95$\downarrow$ & AUROC$\uparrow$ & FPR95$\downarrow$ \\
\midrule
Max log-likelihood & $\novresult{cifar100}{maxLL}{auroc}$ & $\novresult{cifar100}{maxLL}{fpr95}$ & $\novresult{imagenet_r}{maxLL}{auroc}$ & $\novresult{imagenet_r}{maxLL}{fpr95}$ \\
Mahalanobis~\cite{lee2018mahalanobis} & $\novresult{cifar100}{mahalanobis}{auroc}$ & $\novresult{cifar100}{mahalanobis}{fpr95}$ & $\novresult{imagenet_r}{mahalanobis}{auroc}$ & $\novresult{imagenet_r}{mahalanobis}{fpr95}$ \\
Relative Mahalanobis~\cite{ren2021rmd} & $\novresult{cifar100}{rmd}{auroc}$ & $\novresult{cifar100}{rmd}{fpr95}$ & $\novresult{imagenet_r}{rmd}{auroc}$ & $\novresult{imagenet_r}{rmd}{fpr95}$ \\
Energy~\cite{liu2020energy} & $\novresult{cifar100}{energy}{auroc}$ & $\novresult{cifar100}{energy}{fpr95}$ & $\novresult{imagenet_r}{energy}{auroc}$ & $\novresult{imagenet_r}{energy}{fpr95}$ \\
$k$-NN~\cite{sun2022knn} & $\novresult{cifar100}{knn}{auroc}$ & $\novresult{cifar100}{knn}{fpr95}$ & $\novresult{imagenet_r}{knn}{auroc}$ & $\novresult{imagenet_r}{knn}{fpr95}$ \\
\midrule
Competitor ratio $\Lambda$ & $\novresult{cifar100}{lambda}{auroc}$ & $\novresult{cifar100}{lambda}{fpr95}$ & $\novresult{imagenet_r}{lambda}{auroc}$ & $\novresult{imagenet_r}{lambda}{fpr95}$ \\
\bottomrule
\end{tabular}
\vspace{-1.25em}
\end{table*}

\takeaway{A class-moment head uses 1\% of the storage and gains 1–2 pp accuracy over a feature-trained head by exactly solving the balanced problem.}

\subsection{Ablation Study: Sources of Accuracy}
\label{sec:ablation}

\Cref{tab:ablation} varies one component at a time.
The routing and composition blocks share trained adapter weights within each block; the penalty block instead compares independently retrained adapters.

\noindent\textbf{Routing and composition.}
The shared-covariance router raises top-1/top-3 task coverage over per-task mixtures from $83.4/93.5\%$ to $88.0/96.2\%$ on CIFAR-100, from $46.7/65.7\%$ to $69.7/83.5\%$ on ImageNet-R, and from $77.1/92.4\%$ to $89.7/98.1\%$ on CUB-200, using about one-eighth the router storage ($1.49$ versus $12.30$\,MB, packed fp32, CIFAR-100 at $T{=}20$).
With the ridge head, higher coverage improves final accuracy on ImageNet-R ($+0.60{\pm}0.35$\,pp), but not on CIFAR-100 ($-0.15{\pm}0.14$) or CUB-200 ($0.00$); task coverage alone does not determine classification accuracy.
Top-3 activation helps most where coverage is low ($+2.2$\,pp over top-1 on ImageNet-R) and costs $1.1$\,pp on CUB-200, where top-1 routing is already accurate; we keep $k{=}3$ on all benchmarks rather than select it on test data.
In the \emph{GMM + linear head} configuration, weights $1$ instead of $1/3$ cost $1.5$\,pp on CIFAR-100 and $26$\,pp on ImageNet-R, model-sample-calibrated gating adds $0.27{\pm}0.07$\,pp on CIFAR-100, and $K{=}20$ instead of $100$ changes CIFAR-100 accuracy by $-0.05{\pm}0.40$\,pp at a fifth of the router memory.
On CIFAR-100 and ImageNet-R, top-1 classification accuracy exceeds task-identification accuracy: some inputs remain classifiable under a different task's adapter.
A tuned learned-router comparison remains open (Supp.).

\noindent\textbf{Output-factor penalty .}
With retrained adapters and the same seeds, removing $\mathcal L_\perp$
($\lambda_\perp=0$) changes accuracy by $+1.14$ percentage points on CIFAR-100,
$-1.33$ points on ImageNet-R, and $+0.29$ points on CUB-200.

\takeaway{Shared-covariance routing cuts storage ~90\% while improving task retrieval; classification depends on adapter weights (>25 pp loss if poor), while orthogonality changes accuracy only ±1 pp.}


\subsection{Held-out-class novelty detection}
\label{sec:novelty-exp}

\Cref{tab:novelty} uses five class orders: the first 16/20 CIFAR-100
tasks (80 classes) and 8/10 ImageNet-R tasks (160 classes) are known;
remaining classes are novel. Known images are positives. FPR95
uses the fifth percentile of known-test scores, an evaluated ROC
operating point rather than a deployment-calibrated threshold.

Centering raises AUROC over maximum likelihood from $75.6$ to $85.7$
on CIFAR-100 and $60.5$ to $66.0$ on ImageNet-R.
Relative Mahalanobis and linear-probe energy achieve higher AUROC
($88.7/76.8$ and $89.5/74.3$, respectively; CIFAR-100/ImageNet-R)
and lower mean FPR95.
This suggests a benefit from class-level models, although the
comparison does not isolate its cause.
On ImageNet-R, every score has FPR95 near or above $80\%$;
none reliably rejects novel classes at high recall, and $k$-NN
is close to chance ($53.5$ AUROC).
These offline results establish neither online task-arrival
detection nor conformal coverage under drift. Additional protocol
and storage details are in Supp.~\cref{sec:novelty-protocol}.

\takeaway{Centering the router’s existing densities against competing tasks improves unseen-class detection by 5–10 AUROC points over raw likelihood, without changing routing, though dedicated class-level scores perform better.}

\section{Limitations and Conclusion}
\label{sec:conclusion}

\noindent\textbf{Limitations.}
On 10-task ImageNet-R, the fixed-$A$ control outperforms the default.
Pooled covariance updates change earlier task likelihoods. Even with fixed existing scores, new competitors can alter adapter selection when tasks are ambiguous or representations are heavily imbalanced. Therefore, input shifts can invalidate stored statistics despite a frozen encoder.
Inference uses one frozen and one adapted backbone pass; adapter storage is linear in task count and each input needs a frozen routing pass plus an adapted pass—about 48 ms/image on CPU vs. 28 ms/image for a merged model (Supp.~\cref{tab:latency}).
The method, consistent with prior work, assumes a clear separation among tasks and requires that each class be assigned to exactly one task.

\noindent\textbf{Conclusions.}
Class statistics support replay-free routing and classification.
Shared-covariance routing improves task identification over per-task mixtures; ridge heads outperform heads re-balanced on stored features.
\method\ thus treats the geometry of the feature space as memory: A continual learner only needs to retain feature densities, not past task samples.
Routing then becomes retrieval in which a feature vector returns a model rather than another vector, pointing toward adapter libraries that are searched like vector stores and grow by insertion rather than retraining.

\clearpage
\setcounter{section}{0}\setcounter{table}{0}\setcounter{figure}{0}
\setcounter{equation}{0}\setcounter{proposition}{0}\setcounter{lemma}{0}
\setcounter{theorem}{0}\setcounter{remark}{0}\setcounter{algorithm}{0}
\renewcommand{\thesection}{S\arabic{section}}
\renewcommand{\thetable}{S\arabic{table}}
\renewcommand{\thefigure}{S\arabic{figure}}
\renewcommand{\theequation}{S\arabic{equation}}
\renewcommand{\thealgorithm}{S\arabic{algorithm}}
\renewcommand{\theproposition}{S\arabic{proposition}}
\renewcommand{\thelemma}{S\arabic{lemma}}
\renewcommand{\thetheorem}{S\arabic{theorem}}
\renewcommand{\theremark}{S\arabic{remark}}
\def\theHsection{S\arabic{section}}
\def\theHtable{S\arabic{table}}
\def\theHfigure{S\arabic{figure}}
\def\theHequation{S\arabic{equation}}
\def\theHalgorithm{S\arabic{algorithm}}
\def\theHproposition{S\arabic{proposition}}
\def\theHlemma{S\arabic{lemma}}
\def\theHtheorem{S\arabic{theorem}}
\def\theHremark{S\arabic{remark}}
\addtocontents{toc}{\protect\setcounter{tocdepth}{2}}
\twocolumn[{\centering\Large\textbf{\thetitle}\\\vspace{0.5em}Supplementary Material\\\vspace{1.0em}}]
\pdfbookmark[1]{Supplementary Material}{supplementary}

\noindent References of the form ``main Sec.~3.3'', ``main Tab.~2'' or ``main Eq.~(4)'' point to the main paper.
Supplementary sections, tables, figures and equations are prefixed with S.

\tableofcontents

\section{Implementation details}
\label{sec:impl}

\subsection{Pipeline}
\Cref{alg:train,alg:infer} give the training and inference procedures of main Sec.~3.
Completed adapters and recorded class statistics remain fixed. The pooled scatter and ridge moments accumulate, while the shared covariance, ridge head, and training head are updated at each stage. The ridge objective weights classes equally: $G$ sums within-class mean outer products (\cref{sec:ridge-proof}). The router covariance instead averages within-class residual outer products over all seen training samples before shrinkage.
The attention-layer streams have equal numbers of classes per task, so the router may omit the common $-\log|\mathcal{Y}_j|$ term in main Eq.~(2) when ranking tasks; unequal task sizes would require retaining it.
Each class order is trained twice. The replay-free run (main Tab.~2, default) aligns the primary head on Gaussian samples of class statistics and retains no features. The stored-feature run aligns it on stored adapted features and also trains the auxiliary heads of main Tab.~4 on detached features; auxiliary-head operations use separate random generators to avoid changing the primary trajectory.
The ridge head and both routers are evaluated stage by stage from the saved adapters of either run; because adapters are frozen after their stage, this equals computing them during training.
With $k=3$, CIFAR-100's fixed $1/3$ routing weights sum to $1/3$ and $2/3$ at stages 1 and 2. ImageNet-R and CUB-200 normalize the selected routing weights to one. Both conventions coincide from stage 3 onward.

\begin{algorithm}[t]
\small
\caption{Stage $t$ of \method\ (training).}
\label{alg:train}
\begin{algorithmic}[1]
\REQUIRE task data $\mathcal D_t$; frozen backbone; past adapters; router sums; ridge moments; primary head $H$ and its alignment state; auxiliary heads $H_v$ (stored-feature run only)
\STATE initialize $A_t^l\sim\mathcal N(0,1/d)$ and $B_t^l=0$; expand every head to all seen classes
\FOR{each training minibatch $(x,y)$ of task $t$}
  \STATE compute features $h$ using adapter $t$ only, with update $B_t^l A_t^l/r$
  \STATE update adapter $t$ and primary head $H$ using global cross-entropy $+\mathcal L_\perp$
  \STATE update each auxiliary head $H_v$ using cross-entropy on $\operatorname{stopgrad}(h)$
\ENDFOR
\STATE freeze adapter $t$; store frozen-feature class means and add class scatters to $S$ (main Eq.~(2)); fit per-task GMMs for the ablations
\STATE extract current-task adapted features using adapter $t$ at full routing weight; record class means, diagonal variances and counts; add $G_c$ (augmented second moment, averaged within class) to $G$ and store the augmented mean $\mathbf{u}_c$
\STATE replay-free run: discard the adapted features and align $H$ on Gaussian samples of all seen class statistics
\STATE stored-feature run: retain the features and 20-feature subsets, and align $H$ on stored features of all seen classes
\STATE stored-feature run: align sampled and buffer heads on their respective data; leave the unaligned head unchanged
\STATE form $U$ with all seen augmented class means $\mathbf{u}_c$ as columns; solve $W=(G+\lambda I)^{-1}U$; evaluate all heads on the same routed test features
\STATE carry the aligned primary head into stage $t+1$ (its loss trains the next adapter)
\end{algorithmic}
\end{algorithm}

\begin{algorithm}[t]
\small
\caption{Inference after stage $t$.}
\label{alg:infer}
\begin{algorithmic}[1]
\REQUIRE input $\mathbf{x}$; stage $t$; $k$; shared-covariance task densities; frozen adapters; ridge head $W$
\STATE $\LL_j \leftarrow \log \hat p_{j,t}(\phi(\mathbf{x}))$ for $j\le t$ (main Eq.~(2)) \hfill{\footnotesize (one frozen forward pass)}
\STATE $\mathcal{S}\leftarrow$ indices of the $\min(k,t)$ largest $\LL_j$
\STATE $w\leftarrow1/k$ on CIFAR-100, or $1/\min(k,t)$ on ImageNet-R and CUB-200
\STATE $\mathbf{h}\leftarrow$ \texttt{[CLS]} of $f$ with $W^l = W^l_0+(w/r)\sum_{j\in\mathcal{S}}B^l_jA^l_j$ \hfill{\footnotesize (second forward pass)}
\RETURN $\arg\max_c (W^{\top}[\mathbf{h};1])_c$ \hfill{\footnotesize (ridge head over all seen classes)}
\end{algorithmic}
\end{algorithm}

\subsection{Efficient evaluation of the shared-covariance router}
\label{sec:router-evaluation}

At a fixed stage $t$, task $j$ is represented by the
equal-weight mixture
\begin{equation}
\hat p_{j,t}(\mathbf{z})
=
\frac{1}{|\mathcal{Y}_j|}
\sum_{c\in\mathcal{Y}_j}
\mathcal{N}(\mathbf{z}\mid\boldsymbol{\mu}_c,\hat\Sigma_t).
\end{equation}
Assuming $\hat\Sigma_t$ is positive definite, expanding
the Gaussian log-density gives
\begin{equation}
\log\mathcal{N}(\mathbf{z}\mid\boldsymbol{\mu}_c,\hat\Sigma_t)
=
a_{c,t}(\mathbf{z})+b_t(\mathbf{z}),
\end{equation}
where
\begin{align}
a_{c,t}(\mathbf{z})
&=
\boldsymbol{\mu}_c^\top\hat\Sigma_t^{-1}\mathbf{z}
-\frac12
\boldsymbol{\mu}_c^\top\hat\Sigma_t^{-1}\boldsymbol{\mu}_c,
\\
b_t(\mathbf{z})
&=
-\frac12\mathbf{z}^\top\hat\Sigma_t^{-1}\mathbf{z}
-\frac12\log\det(2\pi\hat\Sigma_t).
\end{align}
The class score $a_{c,t}$ is affine in $\mathbf{z}$,
while $b_t$ depends on the input and stage but not on
the class or task.
Factoring $\exp(b_t(\mathbf{z}))$ out of the mixture yields
\begin{equation}
\label{eq:router-logsumexp}
\begin{aligned}
\log\hat p_{j,t}(\mathbf{z})
={}&
\operatorname{LSE}_{c\in\mathcal{Y}_j}
\bigl(a_{c,t}(\mathbf{z})\bigr)
\\
&-\log|\mathcal{Y}_j|+b_t(\mathbf{z}),
\end{aligned}
\end{equation}
where $\operatorname{LSE}_c(a_c)=\log\sum_c\exp(a_c)$.
The affine coefficients can be precomputed after each
covariance update, so scoring requires class-wise affine
evaluations followed by a log-sum-exp within each task.

The term $-\log|\mathcal{Y}_j|$ comes from averaging,
rather than summing, the class densities.
It ensures that each task mixture integrates to one.
For example, if every class density evaluates to the
same value $q$ at an input, the normalized task density
is $q$ regardless of its number of classes; an
unnormalized sum would instead equal $|\mathcal{Y}_j|q$.

For a fixed input, $b_t(\mathbf{z})$ is common to every
task and can be omitted without changing top-$k$ routing.
In the evaluated streams, all tasks also contain the
same number of classes, so $-\log|\mathcal{Y}_j|$ can
likewise be omitted.
For unequal class counts, this normalization must be
retained to rank the defined task densities correctly.
These omissions preserve within-input task rankings;
dropping the input-dependent term $b_t(\mathbf{z})$
does not generally preserve absolute likelihood values
or novelty rankings across different inputs.

\subsection{Hyperparameters}
\label{sec:hyperparameters}
\Cref{tab:hparams} lists the settings of the attention-layer experiments (main Sec.~5).
The ridge coefficient $0.01$ in
$\lambda=0.01\operatorname{tr}(G)/(d+1)$ was selected on held-out
training features of one ImageNet-R class order. Coefficients from
$10^{-4}$ to $1$ produced a held-out accuracy range of $0.22$\,pp on that order.

\begin{table}[t]
\centering
\caption{Hyperparameters of the attention-layer component study (main Tabs.~2, 4 and 5), as in the released code. The additional main Tab.~3 runs are described separately in \cref{sec:published}.}
\label{tab:hparams}
\footnotesize
\setlength{\tabcolsep}{3pt}
\begin{tabularx}{\columnwidth}{@{}lX@{}}
\toprule
Backbone & ViT-B/16, AugReg, ImageNet-21k, frozen \\
LoRA & rank 8, \texttt{attn.qkv}, all 12 blocks; update scaled by $1/r$ \\
Initialization & $A\sim\mathcal{N}(0,1/d)$, $B{=}0$ \\
Optimizer & AdamW, lr $3{\times}10^{-4}$, weight decay $0.01$, cosine decay \\
Batch / epochs per task & 64 / 15 ($T{=}20$), 20 ($T{=}10$) \\
Loss & cross-entropy over all seen classes $+\ \mathcal{L}_\perp$ on $B$, $\lambda_\perp{=}0.1$ \\
Routing density & class means of frozen features, one pooled within-class scatter $S_t$, shrinkage $\gamma{=}0.1$ toward $\operatorname{tr}(S_t/n_t)I/d$, as defined in main Eq.~(2) \\
Ridge head & $W=(G+\lambda I)^{-1}U$ on $[\mathbf{h};1]$, $\lambda=0.01\operatorname{tr}(G)/(d{+}1)$ \\
Ablation GMM & diagonal, $K{=}100$; $k$-means++ init, ${\le}30$ EM iterations, tol.\ $10^{-4}$, variance floor $10^{-6}$; fit uses at most 5,000 current-task features \\
Routing & top-$k$, $k{=}3$, weight $1/3$ (CIFAR-100) or $1/\min(k,t)$ (ImageNet-R, CUB-200) \\
Gate threshold $\tau$ & (ablation) 25th percentile of the top-1/top-2 margin on 100 samples per stored GMM \\
Head re-balancing & SGD, lr $0.01$, momentum $0.9$, weight decay $10^{-4}$, 30 epochs, batch 1024, cosine; $N_c$ samples per class \\
Seeds & CIFAR-100: 0, 7, 42, 1993, 2024; ImageNet-R and CUB-200: 7, 42, 2024 \\
Hardware & NVIDIA L4; per seed at $T{=}20$: \method\ 2.8\,h (5\,h with all head and routing variants), summed controls 4.6--4.8\,h; CUB-200 about 1\,h per run. Total experiments 304.7\,h \\
\bottomrule
\end{tabularx}
\end{table}

\subsection{Final-stage margin-gated ablation}
\label{sec:gate}
This ablation uses the \emph{GMM + linear head} configuration after
all $T$ tasks have been learned. Adapters and the prediction head
remain fixed; only adapter selection and composition change.
It is evaluated once at stage $T$ on all seen tasks, not throughout
training. The default shared-covariance/ridge system does not use it.

We draw 100 synthetic embeddings from each stored task mixture and
score each embedding under all $T$ mixtures. For every embedding,
we subtract its second-highest log-likelihood from its highest.
The threshold is the 25th percentile of these margins pooled across
the synthetic embeddings. A test margin above the threshold selects
the top-1 adapter at full weight; otherwise the top-3 adapters each
receive weight $1/3$. No test examples enter threshold calibration.
At a fixed stage, competitor centering multiplies both the margins
and the matched quantile threshold by $T/(T-1)$, so it leaves these
decisions unchanged.

\subsection{Offline novelty protocol}
\label{sec:novelty-protocol}
Main Tab.~6 evaluates frozen-feature scores on five class orders.
The first 16 of 20 CIFAR-100 tasks (80 classes), or 8 of 10
ImageNet-R tasks (160 classes), supply known classes; test images
from the remaining classes are novel. Known images are positives.
FPR95 is the fraction of novel inputs accepted at the fifth
percentile of known-test scores. This is a test ROC operating
point, not a threshold calibrated for deployment.

Max-LL and competitor centering use per-task GMMs. Mahalanobis uses
class means and a tied full covariance; relative Mahalanobis also
fits a global mean and covariance. These frozen-feature matrices
differ from the diagonal adapted-feature variances used for head
alignment and incur additional fitting and storage. Energy uses a
linear probe trained on known classes. The $k$-NN reference uses
$k=50$ and retains every known training feature.
No empirical conformal transformation is applied; such transforms
can introduce ties and need not preserve the raw-score AUROC/FPR95.

\subsection{Baselines and matched protocol}
\label{sec:baselines}
The results labeled \emph{Sum, fixed-$A$} and \emph{Sum, trainable-$B$} in the paper are local summed-adapter controls.
In our setup, these controls are produced by running the project runner (e.g., a CLI entry point) with the corresponding method flags (\texttt{sum\_fixed\_a} and \texttt{sum\_trainable\_b}).
The fixed-$A$ control draws an independent Gaussian matrix per task and block, uses its QR factor as $A_t$, trains $B_t$, then freezes both. It captures the fixed-basis idea of BiLoRA~\cite{zhu2025bilora} with one fixed input basis, without BiLoRA's two-basis bilinear construction.
The trainable-$B$ control freezes past $A$ factors but updates past $B$ factors. It captures SD-LoRA's idea of keeping earlier directions fixed~\cite{wu2025sdlora}, without SD-LoRA's fixed-direction/learned-magnitude parameterization.
The published methods themselves are evaluated using their official implementations, rather than via these local controls.

SLAO is a local vision adaptation of the merging rule in~\cite{qiao2025slao}, whose original experiments use language models.
It QR-orthonormalizes the previous $A$ row space to initialize the new $A$, and blends $B$ with weight $1/\sqrt t$ while replacing $A$ with its new value. Reduced QR spans the previous row space; it does not construct its orthogonal complement.
Because this initialization departs from the published rule and the corresponding archived training script is not part of the public release, the main paper does not use SLAO as a comparator. For completeness, the five archived CIFAR-100 runs reach $83.04{\pm}1.14\%$ final accuracy and $14.12{\pm}0.99\%$ forgetting; ImageNet-R was not run.

The CIFAR summed-control rows combine archived A5 runs for seeds 0, 7, 42, and 2024 with new seed-1993 runs generated with the current runner; the old seed-1993 runs are excluded because their head-evaluation protocol differed.
The new seed-1993 accuracies are $83.64\%$ (fixed-$A$) and $34.85\%$ (trainable-$B$).
The latter is below the other trainable-$B$ seeds, not within their range.
The \method\ and ImageNet-R records underlying main Tab.~2 use the current runner.
As a check that the current runner reproduces the archived protocol, \method\ was trained by both on seeds 0, 7, 42, and 2024: final accuracies differ by at most $0.6$\,pp per seed ($86.49\%$ archived versus $86.48\%$ current, mean of four). The two summed controls share this port, but their own archived seed-1993 runs ended near $9\%$, so no same-seed check exists for them.
The shared class-order seeds support paired descriptive comparisons; they do not remove differences in implementation, random-number consumption, or training policy.

The comparisons in main Tab.~3 (L2P, InfLoRA, SD-LoRA, C-LoRA, CL-LoRA) are separate from these local controls; each method is trained with its official code. BiLoRA and matched HESTIA comparisons are not included in Tab.~3.

\noindent\textbf{Training details affecting interpretation.}
Both training and evaluation use the deterministic timm evaluation transform; the backbone remains in evaluation mode while adapter gradients are enabled.
Head losses use a global softmax over all seen classes, including when the minibatch contains only current-task labels.
Alignment uses each class's original feature count $N_c$. Thus it restores access to past classes but does not equalize class frequencies on ImageNet-R.
The buffer head instead uses at most 20 features per class.
Mean/variance sampling is performed once per alignment stage, then reused across its 30 SGD epochs.
The original projection of the zero-initialized $B_t$ was a no-op and is absent from the current runner; the evaluated regularization is the $B_t^\top B_s$ penalty.

\section{Proofs}
\label{sec:proofs}

\subsection{Ridge head from empirical class moments}
\label{sec:ridge-proof}
At a fixed stage, let $C$ classes have augmented training features $\tilde h_{c,a}$, $a=1,\dots,N_c$, computed using their own frozen adapters at full weight. Define $G_c=N_c^{-1}\sum_a\tilde h_{c,a}\tilde h_{c,a}^{\top}$, $u_c=N_c^{-1}\sum_a\tilde h_{c,a}$, $G=\sum_cG_c$ and $U=[u_1,\dots,u_C]$.
For one-hot targets $e_c$, the implemented objective is
\begin{equation}
\label{eq:ridge-objective}
J(W)=\sum_{c=1}^{C}\frac{1}{N_c}\sum_{a=1}^{N_c}
\|W^{\top}\tilde h_{c,a}-e_c\|^2+\lambda\|W\|_F^2.
\end{equation}
Expanding gives $J(W)=\operatorname{tr}(W^{\top}GW)-2\operatorname{tr}(W^{\top}U)+C+\lambda\|W\|_F^2$.
Since $G\succeq0$ and $\lambda>0$, setting the gradient to zero gives the unique solution $W=(G+\lambda I)^{-1}U$.
This is an exact sufficient-statistic calculation for the fixed empirical features, up to numerical precision; it does not require Gaussian features. Each class contributes equally, irrespective of $N_c$.
If the data term is instead averaged over classes, its regularizer must be $\lambda/C$ to give the same solution. The identity matrix also regularizes the augmented bias coordinate.
The feature map used by the test-time adapter composition differs from the own-adapter map used to collect these moments, so replay equivalence for this objective is not a classification guarantee.

\subsection{Routing error (main Prop.~1) and its \texorpdfstring{top-$k$}{top-k} version}
Fix a stage $t$ and abbreviate $\hat p_{j,t}$ as $\hat p_j$. These are Lebesgue densities on the frozen embedding space. Let $\hat P_i$ have density $\hat p_i$, $P_i$ be the true distribution of $\phi(\mathbf{x})$ for task $i$, and $\mathrm{OVL}(\hat p_i,\hat p_j)=\int_{\mathbb R^d}\min\{\hat p_i(\mathbf{z}),\hat p_j(\mathbf{z})\}\,d\mathbf{z}$.
All probabilities are conditional on the fitted descriptors, with the test input independent of the fitting data.
Write $E_j=\{\mathbf{z}:\hat p_j(\mathbf{z})\ge\hat p_i(\mathbf{z})\}$; each $E_j$ is measurable because mixture densities are continuous.
For a fixed event $E$ let $\delta_i(E)=|P_i(E)-\hat P_i(E)|$; with $\mathrm{TV}(P,Q)=\sup_A|P(A)-Q(A)|$, $\delta_i(E)\le\mathrm{TV}(P_i,\hat P_i)$.

\begin{lemma}
\label{lem:pair}
$\hat P_i(E_j)\le\mathrm{OVL}(\hat p_i,\hat p_j)$.
\end{lemma}
\begin{proof}
On $E_j$, $\hat p_i=\min(\hat p_i,\hat p_j)$, so $\hat P_i(E_j)=\int_{E_j}\min(\hat p_i,\hat p_j)\le\int\min(\hat p_i,\hat p_j)$.
\end{proof}

\begin{proposition}[main Prop.~1]
Top-1 routing errs on task $i$ with $P_i$-probability at most $\sum_{j\ne i}\mathrm{OVL}(\hat p_i,\hat p_j)+\delta_i(E)$, where $E=\bigcup_{j\ne i}E_j$.
\end{proposition}
\begin{proof}
Under any tie-breaking rule the error event is contained in $E$.
Then $P_i(E)\le\hat P_i(E)+\delta_i(E)$, and a union bound with \cref{lem:pair} bounds $\hat P_i(E)$.
\end{proof}

\begin{proposition}[Top-$k$ routing]
\label{prop:topk}
For $k<t$, the probability that task $i$ is not among the $k$ highest-scoring tasks is at most $\frac{1}{k}\sum_{j\ne i}\mathrm{OVL}(\hat p_i,\hat p_j)+\delta_i(\{N\ge k\})$, with $N(\mathbf{z})=\sum_{j\ne i}\mathbf{1}[\mathbf{z}\in E_j]$. For $k\ge t$ the event is empty.
\end{proposition}
\begin{proof}
Under any tie-breaking rule, task $i$ is missed only if $N(\mathbf{z})\ge k$.
Under $\hat P_i$, Markov's inequality and \cref{lem:pair} give $\hat P_i(N\ge k)\le\frac1k\sum_{j\ne i}\hat P_i(E_j)\le\frac1k\sum_{j\ne i}\mathrm{OVL}(\hat p_i,\hat p_j)$; transfer to $P_i$ as before.
\end{proof}
\noindent\textbf{Gated selection.}
Let $Q$ be the event that the gate selects top-1, with its score function and threshold fixed before observing the test input. A missed true task is contained in
$H=(Q\cap\{N\ge1\})\cup(Q^c\cap\{N\ge k\})$.
Pointwise, $\mathbf{1}_H\le(\mathbf{1}_Q+\mathbf{1}_{Q^c}/k)N$, so
\begin{equation}
\begin{aligned}
P_i(\mathrm{miss})\le{}&\sum_{j\ne i}\int_{E_j}
\left(\mathbf{1}_Q+\tfrac{\mathbf{1}_{Q^c}}{k}\right)
\min(\hat p_i,\hat p_j)\\
&+\delta_i(H).
\end{aligned}
\end{equation}
This follows by integration under $\hat P_i$ and transfer on the single event $H$. The unconditional top-1 and top-$k$ bounds cannot simply be conditioned on the gate: that changes the input distribution.

\noindent\textbf{Scope.}
The overlaps involve only stored descriptors and can be estimated by Monte Carlo from them.
The fit term is the reason we use $\delta_i$ rather than total variation.
Under our deterministic evaluation transform, the discrete distribution
of 8-bit images induces an atomic $P_i$. Its countable support has
$P_i$-mass 1 and $\hat P_i$-mass 0 because $\hat P_i$ has a Lebesgue
density. Thus $\mathrm{TV}(P_i,\hat P_i)=1$; this support mismatch
alone does not imply poor routing. Continuous stochastic preprocessing
would require reassessing the atomic-support premise.
At a fixed stage, estimate $P_i(E)$ by the empirical frequency of $E$
on held-out task-$i$ inputs, and $\hat P_i(E)$ by its frequency on
independent samples from the fitted model. Their absolute difference
estimates $\delta_i(E)$. A risk certificate additionally requires
sampling-error control for both estimates (and estimated overlaps);
we do not report such a certificate.
For independently fitted, frozen per-task mixtures, adding a task leaves old pairwise overlaps fixed and adds a new competitor. The main shared-covariance router instead updates $\hat\Sigma_t$: old densities, their overlaps and the error event all change. The bounds apply anew at each stage, without an additive-across-stages guarantee.
The bounds concern routing on frozen features only; they say nothing about the adapted features used by the classifier.
Their fixed-fit interpretation assumes the true task distribution does not change. A frozen representation alone does not prevent covariate shift.

\subsection{Variance of the competitor ratio (main Lemma~1)}
For $t>1$, assume $\Var(\LL_j)=\sigma^2>0$ for all $j$ and $\Cov(\LL_j,\LL_q)=\rho\sigma^2$ for $j\ne q$, and let $\overline{\LL}_{-j}=\frac{1}{t-1}\sum_{q\ne j}\LL_q$.
This common-correlation covariance requires $-1/(t-1)\le\rho\le1$.
Then
\begin{align}
\Var(\overline{\LL}_{-j}) &= \tfrac{\sigma^2}{(t-1)^2}\big[(t-1)+(t-1)(t-2)\rho\big]\nonumber\\
 &= \tfrac{\sigma^2\,[1+(t-2)\rho]}{t-1},\nonumber\\
\Cov(\LL_j,\overline{\LL}_{-j}) &= \rho\sigma^2,\nonumber
\end{align}
and therefore
\begin{equation}
\Var(\Lambda_j)=\sigma^2-2\rho\sigma^2+\tfrac{\sigma^2[1+(t-2)\rho]}{t-1}=\sigma^2\,\tfrac{t(1-\rho)}{t-1}.
\end{equation}
The random variable is the input drawn from a specified reference population, with fitted models held fixed. Equal variances and equicorrelation are assumptions, not consequences of high dimension; Gaussian scores are not required for the identity.
The lemma concerns variance only; detection performance must be evaluated independently. The next subsection explains why the historical SNR statistic is insufficient for that purpose.

\subsection{What the historical SNR statistic measures}
\label{sec:snr}
The measurement code computes, for each labeled input, $a=\LL_y$ and $b=(t-1)^{-1}\sum_{j\ne y}\LL_j$, with $y$ the true task, and $D=a-b$.
It reports
\begin{equation}
S_{\LL}=\frac{\mathbb E[a]-\mathbb E[b]}{\operatorname{sd}(b)}.
\end{equation}
After centering, the true-task score is $D$ and the mean of the wrong-task scores is $-D/(t-1)$ because the centered scores sum to zero.
Ignoring the code's numerical $10^{-8}$ stabilizer, and assuming nonzero standard deviations and $\mathbb E[D]\ne0$ (the ratio divides by $S_{\LL}$),
\begin{equation}
S_\Lambda=t\frac{\mathbb E[D]}{\operatorname{sd}(D)},\qquad
\frac{S_\Lambda}{S_{\LL}}=t\frac{\operatorname{sd}(b)}{\operatorname{sd}(D)}.
\end{equation}
This is a task-labeled score statistic, not known-versus-novel separation.
Its ratio includes a task-count factor introduced by changing the denominator; it cannot be interpreted as a like-for-like variance reduction or a detection gain.
The previous runner-up/competitor-mean decomposition does not match the code and is not used here. We report direct held-out-class detection metrics in the main paper instead.

\subsection{Conditions for a conformal novelty construction}
\label{sec:conformal}
No conformal detector is executed by the reported novelty script; the following describes a possible construction and its assumptions.
Fix the fitted models and the current task set before calibration.
For task $j$, obtain calibration points $Z_{j1},\dots,Z_{jm}$ independently of fitting, and define
\begin{equation}
\pi_j(x)=\frac{1+\sum_{i=1}^m\mathbf{1}\{\Lambda_j(Z_{ji})\le\Lambda_j(x)\}}{m+1}.
\end{equation}
If the calibration points and a fresh task-$j$ test point are exchangeable under the same score function, then $\Pr[\pi_j(X)\le\alpha]\le\alpha$~\cite{vovk2005algorithmic}.
With ties the rank need not be uniform, but this conservative inequality holds.
When task identity is unknown, $\pi_{\mathrm{known}}(x)=\max_j \pi_j(x)$ provides a conservative known-class rejection rule: for the true task $i$, $\{\pi_{\mathrm{known}}\le\alpha\}\subseteq\{\pi_i\le\alpha\}$.
Using only the likelihood-selected task's $p$-value does not have this implication; a Bonferroni factor alone does not repair testing the wrong task model.

Adding competitors changes $\Lambda$, requiring calibration under the new score function.
Resampling from a fitted GMM gives a guarantee only for a fresh test point from that same fitted model, not for real task data unless the fit is exact.
If instead calibration points are drawn from $\hat P_j$ and the test point from the real task distribution $P_j$, the level becomes $\alpha+\mathrm{TV}(P_j,\hat P_j)$, a valid but loose upper bound; for the atomic $P_j$ of 8-bit images this distance is 1, so the transfer is vacuous.
No such error term or online false-rejection rate is measured here.
Finite empirical rank transforms also introduce ties, so AUROC and FPR95 need not equal those of continuous raw scores.

\subsection{Why factor orthogonality is not a forgetting guarantee}
\label{sec:forgetting}
The implemented penalty controls $\|B_t^\top B_s\|_F^2$.
Although
\begin{equation}
|\langle B_tA_t,B_sA_s\rangle_F|
\le \|B_t^\top B_s\|_F\,\|A_tA_s^\top\|_F,
\end{equation}
a small parameter inner product does not bound the output perturbation on an old input.
For example, take rank one, $A_s=e_1^\top$, $A_t=e_2^\top$, $B_s=e_1$, and $B_t=e_2$.
Both factor overlaps are zero, so the penalty and the parameter inner product vanish, but on the old-task input $e_1+e_2$ the new adapter adds $w\,e_2/r$ to the output under main Eq.~(3). A downstream classifier sensitive to that direction can change its prediction.
More generally, a nonzero update $B_tA_t$ perturbs every input outside its null space, even when its factors are orthogonal to those of past adapters.
Thus the former proof step that bounded output change by alignment with a past update was unsupported.
Moreover, gradient descent on factors with this penalty is not gradient descent on normalized full-update alignment coefficients.
We do not claim an $O(t^{-1/2})$ forgetting law, nor use a fitted decay curve as validation of one.
The no-penalty retraining experiment is the evidence for the regularizer's effect in this implementation.

\section{Additional results}
\label{sec:results}

\subsection{Per-task accuracy at \texorpdfstring{$T{=}20$}{T=20}}
\Cref{tab:pertask} breaks main Tab.~2 down by task.
The replay-free \method\ leads on the eight oldest tasks; the fixed-$A$ control, which re-balances on stored features, leads on tasks 8--18. These are descriptive differences across the evaluated configurations.
The trainable-$B$ control reaches $98.88\%$ on the last task and stays below $67\%$ on every task before task 18.

\begin{table}[t]
\centering
\caption{Accuracy on task $j$ after the last stage, CIFAR-100, $T{=}20$, mean $\pm$ std over the same five seeds for all methods. \method: replay-free, shared-covariance router, ridge head.}
\label{tab:pertask}
\scriptsize
\setlength{\tabcolsep}{2pt}
\begin{tabular}{c ccc}
\toprule
$j$ & Sum, trainable-$B$ & Sum, fixed-$A$ & \method\ \\
\midrule
 0 & $32.84{\pm}21.27$ & $55.48{\pm}12.96$ & $\mathbf{85.68{\pm}4.93}$ \\
 1 & $29.52{\pm}26.32$ & $59.64{\pm}13.07$ & $\mathbf{87.80{\pm}3.36}$ \\
 2 & $31.96{\pm}12.72$ & $54.56{\pm}14.83$ & $\mathbf{84.24{\pm}6.25}$ \\
 3 & $33.96{\pm}18.86$ & $68.84{\pm}7.44$ & $\mathbf{87.36{\pm}4.08}$ \\
 4 & $31.04{\pm}15.29$ & $77.44{\pm}8.48$ & $\mathbf{88.92{\pm}2.26}$ \\
 5 & $38.40{\pm}15.46$ & $81.76{\pm}6.92$ & $\mathbf{88.76{\pm}3.40}$ \\
 6 & $36.08{\pm}14.84$ & $82.32{\pm}2.92$ & $\mathbf{86.60{\pm}3.28}$ \\
 7 & $45.12{\pm}23.12$ & $90.24{\pm}3.87$ & $\mathbf{90.72{\pm}2.39}$ \\
 8 & $49.08{\pm}13.84$ & $\mathbf{91.88{\pm}2.13}$ & $89.88{\pm}4.73$ \\
 9 & $53.72{\pm}11.08$ & $\mathbf{91.32{\pm}2.17}$ & $89.92{\pm}1.43$ \\
10 & $59.00{\pm}19.17$ & $\mathbf{92.72{\pm}3.59}$ & $91.08{\pm}2.30$ \\
11 & $41.76{\pm}13.81$ & $\mathbf{90.64{\pm}4.66}$ & $88.32{\pm}3.68$ \\
12 & $41.40{\pm}14.32$ & $\mathbf{91.20{\pm}4.44}$ & $87.16{\pm}3.36$ \\
13 & $56.12{\pm}21.16$ & $\mathbf{93.08{\pm}3.86}$ & $89.68{\pm}2.52$ \\
14 & $60.80{\pm}17.11$ & $\mathbf{93.84{\pm}2.23}$ & $90.36{\pm}2.09$ \\
15 & $64.60{\pm}13.41$ & $\mathbf{94.48{\pm}1.83}$ & $89.60{\pm}2.73$ \\
16 & $58.32{\pm}11.60$ & $\mathbf{94.76{\pm}0.64}$ & $89.44{\pm}2.15$ \\
17 & $66.44{\pm}12.13$ & $\mathbf{94.96{\pm}1.66}$ & $88.32{\pm}2.08$ \\
18 & $86.68{\pm}13.73$ & $\mathbf{96.00{\pm}1.33}$ & $87.76{\pm}2.54$ \\
19 & $\mathbf{98.88{\pm}0.58}$ & $95.80{\pm}1.77$ & $89.48{\pm}3.43$ \\
\midrule
Mean & 50.79 & 84.55 & \textbf{88.55} \\
Fgt  & 50.80 & 12.88 & \textbf{5.16} \\
\bottomrule
\end{tabular}
\end{table}

\subsection{Ten tasks, nearest-class-mean evaluation}
\label{sec:t10}
Before the $T{=}20$ study we ran CIFAR-100 with 10 tasks of 10 classes (seed 1993, 20 epochs per task), classifying by nearest class mean computed from 20 stored adapted features per class, except for the explicitly labeled linear-head row.
These historical rows differ in head and routing settings and are not used as a controlled comparison on one verified checkpoint.
The base configuration ($K{=}20$, top-1) uses NCM. The $K{=}100$, top-3 result uses a calibrated linear head, as confirmed by the experiment log; their difference is not a routing-only ablation.
As predicted by main Sec.~3.5, the competitor ratio does not change the result when routing is a per-input arg-max: the GMM and GMM+$\Lambda$ rows are identical.
Only the current adapter is active during \method\ training; the summed controls run all adapters in the forward pass. In these historical runs, summed-control task times grew from $19.5$ to $34.3$\,min by task 10; total times were 3.3\,h for \method\ versus 4.4--4.5\,h for the controls. This does not imply constant total cost per task: classifier training and alignment also depend on the number of seen classes.

\begin{table}[t]
\centering
\caption{Historical CIFAR-100, 10 tasks, seed 1993. NCM except $^\ast$, which uses a calibrated linear head. Std is spread across tasks, not uncertainty across seeds.}
\label{tab:t10}
\footnotesize
\setlength{\tabcolsep}{2.5pt}
\begin{tabular}{lccc}
\toprule
\textbf{Method} & Acc & Std & Train (h) \\
\midrule
Sum, trainable-$B$ & 62.64 & 17.4 & 4.5 \\
Sum, fixed-$A$ & 85.50 & 8.2 & 4.4 \\
\midrule
\method, $K{=}20$, top-1 & 83.55 & 3.1 & 3.3 \\
\method, $K{=}20$, top-1, $\Lambda$ & 83.55 & 3.1 & 3.3 \\
\method, $K{=}100$, top-3$^\ast$ & \textbf{87.50} & 2.75 & 3.3 \\
\bottomrule
\end{tabular}
\end{table}

\subsection{Routing diagnostics and missing controls}
\label{sec:routers}
The reported routing comparisons use the same layer-level composition and trained adapters.
A tuned learned-router comparison under the same training and evaluation protocol remains missing.
Historical explorations used cached feature mixtures and test-derived tuning data; they are excluded from comparative evidence here.
A valid comparison must train on permitted stage data, tune on a separate validation split, and evaluate layer-level composition on untouched test inputs.

\begin{figure}[t]
\centering
\includegraphics[width=\columnwidth]{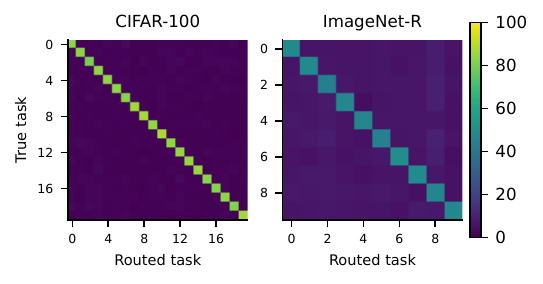}
\caption{Top-1 routing confusion at the end of the stream (rows: true task, columns: routed task, row-normalized, mean over seeds; $K{=}100$). CIFAR-100 ($T{=}20$, 5 seeds) is nearly diagonal; on ImageNet-R ($T{=}10$, 3 seeds) errors spread over all tasks.}
\label{fig:confusion}
\end{figure}

\begin{figure}[t]
\centering
\includegraphics[width=0.9\columnwidth]{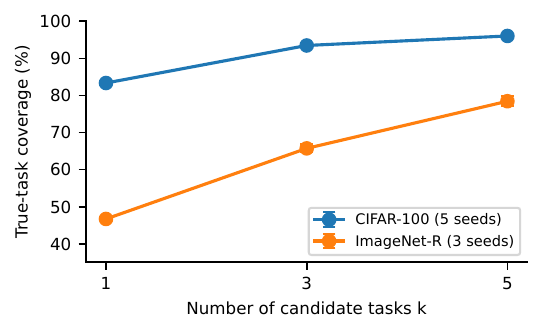}
\caption{Routing coverage: fraction of test inputs whose true task is among the $k$ highest-likelihood tasks, measured at $k=1,3,5$ in the final runs. CIFAR-100: five seeds; ImageNet-R: three; 95\% Student-$t$ intervals.}
\label{fig:kmax}
\end{figure}

\noindent\textbf{Routing coverage at the end of the stream.}
\Cref{tab:coverage} compares the two task descriptors on the replay-free runs.
Both read frozen features, so coverage does not depend on the adapters.
\Cref{fig:confusion,fig:kmax} show the per-task GMM of the \emph{GMM + linear head} configuration.

\begin{table}[t]
\centering
\caption{Top-1 / top-3 task coverage (\%) at the final stage, mean over class orders. Router storage uses packed fp32 arrays at the largest task count; each pair lists per-task GMM / shared covariance.}
\label{tab:coverage}
\footnotesize
\setlength{\tabcolsep}{3pt}
\begin{tabular}{lccc}
\toprule
\textbf{Descriptor} & \textbf{CIFAR-100} & \textbf{ImageNet-R} & \textbf{CUB-200} \\
\midrule
Per-task GMM, $K{=}100$ & 83.4 / 93.5 & 46.7 / 65.7 & 77.1 / 92.4 \\
Class means, shared cov. & 88.0 / 96.2 & 69.7 / 83.5 & 89.7 / 98.1 \\
\midrule
Memory (MB) & 12.3 / 1.49 & 6.15 / 1.80 & 6.15 / 1.80 \\
\bottomrule
\end{tabular}
\end{table}

\noindent\textbf{Per-task mixture diagnostics.}
The fitted $K{=}100$ mixtures show signs of overfitting: several components concentrate on few points, and training log-likelihood exceeds held-out log-likelihood by roughly 70 nats on CIFAR-100 and ImageNet-R. A diagonal mixture can model correlations through its component means, but its within-component covariances remain diagonal.
Removing the log-determinant term lowers ImageNet-R top-1 coverage to $26.2\%$; this correction does not resolve the routing errors, but the test does not rule out other score corrections.
A per-class diagonal model reaches $87.6\%$ top-1 on CUB-200; a shared full covariance additionally represents within-class correlations using one matrix for all tasks.
With shrinkage $\gamma\in\{0.01,0.1,0.3\}$ ImageNet-R top-1 coverage varies by less than $0.4$\,pp.

\noindent\textbf{Two attempts that did not help (ImageNet-R).}
Both are post-training evaluations on fixed adapters from stored-feature-training runs, using the ridge head; neither is a separate replay-free training experiment.
(i)~\emph{Moment-based weights.} Instead of equal weights over the top-3 adapters, we solved a simplex-constrained quadratic program that uses per-layer activation second moments to penalize deviations from individual adapter updates. On class order 7 it gave $71.85\%$, close to posterior weights ($71.83\%$) and below equal weights ($73.00\%$).
(ii)~\emph{Shortlist verification.} The router shortlists $k$ tasks, each runs its own adapter at full weight, and the ridge head scores that task's classes; the highest score wins. Final accuracy was $70.79$ / $72.27\%$ ($k{=}3$ / $5$) on order 7 and $71.32$ / $72.56\%$ on order 2024, against $72.98$ and $73.67\%$ for top-3 composition; this scoring rule did not improve accuracy on either order.
A local HESTIA-inspired diagnostic~\cite{le2026taskfree}, not a reproduction, evaluates fixed adapters from a stored-feature-training run using adapted-feature mixtures with one pass per task. On order 7, it gave $68.08\%$ final accuracy with the stored-feature head, against $69.82\%$ for top-1 routing with the shared-covariance density.
Coverage is image-weighted; the downstream classification metric averages tasks equally.
\Cref{fig:confusion,fig:kmax} are regenerated from the final-run JSONs.
Coverage says whether the true task is available to the composition, not whether that composition predicts correctly.

\subsection{Reference points}
The CIFAR-100 and ImageNet-R joint-training references in main Tab.~2 use rank-32 LoRA and 3 epochs on all classes jointly: $92.90{\pm}0.14\%$ on CIFAR-100 (five archived seeds) and $81.69\%$ on ImageNet-R (one new seed).
The completed CUB-200 joint experiment gives $87.79{\pm}0.16\%$, $0.80$\,pp above the default continual result.
These are separate training references, not upper bounds. Oracle selection additionally receives task identity and activates one adapter at full weight, whereas the default composes three; the oracle gap therefore combines selection and composition effects.

\subsection{Additional comparison protocol}
\label{sec:published}
Main Tab.~3 reports an additional set of comparison runs, with 20 tasks of 5 classes on CIFAR-100, 25 tasks of 8 classes on ImageNet-R, and 10 tasks of 20 classes on CUB-200.
Its ImageNet-R stream therefore differs from the 10-task stream used for the component analysis in main Tabs.~2, 4 and 5 and for the held-out-class novelty study.
The \method\ means in main Tab.~3 also come from these additional runs: $88.52/88.69$, $71.92/71.87$ and $86.86/87.12\%$ for replay-free/stored-feature training on the three benchmarks, respectively.
They are not replacements for the matched class-order means and intervals in main Tab.~2.

The table reports additional experiments with L2P, InfLoRA, SD-LoRA, C-LoRA and CL-LoRA; citations identify the algorithms.
Backbone checkpoints, class orders and training budgets are method-specific, so the table provides descriptive comparisons rather than a paired component ablation.
It reports aggregate final accuracies without per-seed uncertainty; we make no significance claim from these means.

\subsection{Geometric descriptor variants}
\label{sec:extended}
\begin{figure*}[t]
\centering
\includegraphics[width=\textwidth]{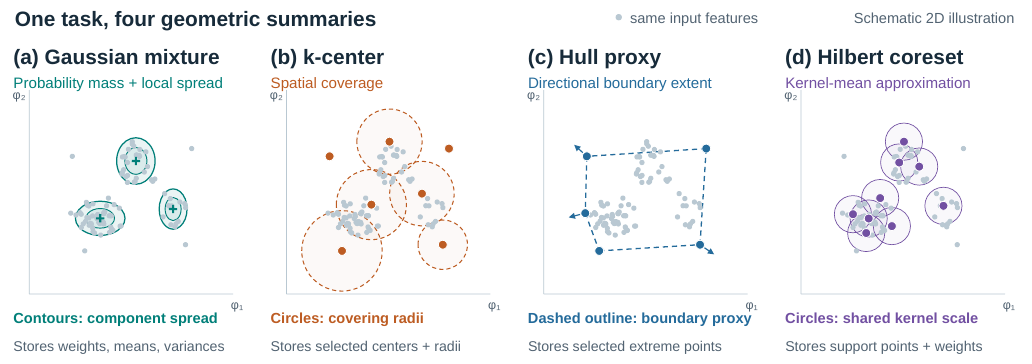}
\caption{Four geometric descriptors of the same schematic task. Gray points are input features. (a) A GMM stores component weights, means (crosses), and diagonal variances (contours). (b) $k$-center retains centers and covering radii. (c) The hull proxy retains points extreme along sampled directions; dashed edges illustrate their boundary envelope, not an exact hull-distance score. (d) A Hilbert coreset selects support points to approximate the kernel mean; equal-radius circles indicate the shared kernel scale, not covering radii. Panels (b--d) retain selected feature vectors. Budgets are reduced for illustration; this is not a benchmark projection.}
\label{fig:descriptor-geometry}
\end{figure*}

\Cref{tab:selection-primary,tab:selection-extended} compare four ways to summarize task geometry: probability density, spatial coverage, boundary extent, and kernel mean embeddings.
\Cref{fig:descriptor-geometry} illustrates these distinct representations on the same schematic task.
All variants use frozen ViT-B/16 \texttt{[CLS]} features in a known-vs-novel selection study, without attention-layer adapters or the ridge prediction head.

\noindent\textbf{Gaussian mixtures (GMM).}
A diagonal mixture fitted by EM~\cite{dempster1977em} stores component weights, means, and coordinate-wise variances (\descriptorref{a}).
The means locate components, the variances describe their local spread, and the weights allocate probability mass; the mixture log-density scores familiarity with a task.
Each task uses $K{=}20$ components.

\noindent\textbf{$k$-center.}
Farthest-first selection retains $50$ feature vectors as centers, together with their covering radii (\descriptorref{b}).
Each new center covers a region far from those already selected; negative nearest-center distance supplies a familiarity score.
This descriptor emphasizes spatial coverage rather than modeling how probability mass is distributed within the covered regions.

\noindent\textbf{Convex-hull proxy.}
Directional-extrema selection retains feature vectors that maximize projections along sampled directions (\descriptorref{c}).
These points summarize the task's outer extent, emphasizing boundary geometry rather than interior density.
The descriptor is a finite-direction proxy, not an exact convex hull or an exact point-to-hull distance computation.

\noindent\textbf{Hilbert coreset.}
Kernel-herding-style selection~\cite{chen2010herding} retains $40$ support vectors to approximate the empirical mean in a Gaussian-kernel feature space (\descriptorref{d}), using random Fourier features~\cite{rahimi2007random} during selection.
The score is the log of the weighted Gaussian-kernel similarity to these vectors, with uniform weights.
It summarizes distributional geometry at the kernel bandwidth's scale; this similarity score is not a normalized density.
Unlike the parametric GMM and the default statistical-memory system, the $k$-center, hull-proxy, and Hilbert variants retain selected per-example feature vectors.

\noindent\textbf{Projection and score calibration.}
+JL constructs the descriptor after a shared Johnson--Lindenstrauss projection to $d'{=}256$, reducing its feature dimension.
+F denotes competitor-centered scoring with conformal threshold calibration in this study.
Centering changes cross-input threshold decisions while preserving the per-input task ranking; it does not change which geometry the descriptor stores.

\noindent\textbf{Evaluation.}
Known tasks are evaluated alongside held-out novel-task probes.
The reported known-task counts are $T{=}19$ for CIFAR-100 and ImageNet-R, $T{=}9$ for ImageNet-A, CUB-200 and Cars, $T{=}5$ for VTAB-50, and $T{=}4$ for DomainNet.
$A$ is balanced known-vs-novel selection accuracy, with a $50\%$ chance reference; $A_T$ is downstream accuracy after the final task.
The feature-only model, held-out-task protocol, and metrics differ from main Tabs.~2 and~3, so their accuracy levels are not directly comparable.
For the unprojected GMM, adding +F changes $A_T$ from $51.73$ to $62.85\%$ on VTAB-50 and $79.50$ to $83.51\%$ on CUB-200, and selection accuracy from $44.1$ to $54.4\%$ on DomainNet. Its $A_T$ changes on CIFAR-100 and ImageNet-R are $+0.74$ and $-0.75$\,pp. Effects depend on the descriptor: Hilbert+F lowers VTAB-50 $A_T$ from $59.84$ to $54.21\%$.
The observed change with +F also depends on projection: for the ImageNet-A GMM, $A_T$ rises from $58.44$ to $60.15\%$ without JL, but falls from $59.37$ to $52.23\%$ with JL.
The tables report no per-seed uncertainty; these comparisons are descriptive and concern the thresholded protocol, not changes to per-input density arg-max routing.

\begin{table}[t]
\centering
\caption{Geometric descriptor variants for known-vs-novel selection (\cref{sec:extended}), primary benchmarks. $A$: balanced selection accuracy; $A_T$: downstream accuracy after the final task (\%). \textbf{Bold}: best variant. Mem: per-stage descriptor memory, CIFAR-100 configuration.}
\label{tab:selection-primary}
\scriptsize
\setlength{\tabcolsep}{1.1pt}
\resizebox{\columnwidth}{!}{%
\begin{tabular}{l cc cc cc cc r}
\toprule
& \multicolumn{2}{c}{\textbf{C100} ($T{=}19$)} & \multicolumn{2}{c}{\textbf{IN-R} ($T{=}19$)} & \multicolumn{2}{c}{\textbf{IN-A} ($T{=}9$)} & \multicolumn{2}{c}{\textbf{VTAB} ($T{=}5$)} & \\
\cmidrule(lr){2-3}\cmidrule(lr){4-5}\cmidrule(lr){6-7}\cmidrule(lr){8-9}
\textbf{Variant} & $A_T$ & $A$ & $A_T$ & $A$ & $A_T$ & $A$ & $A_T$ & $A$ & Mem \\
\midrule
\gmm & 48.17 & 56.6 & \textbf{51.08} & \textbf{58.9} & 58.44 & 64.2 & 51.73 & 58.1 & 120\,KB \\
\gmmf & \textbf{48.91} & \textbf{57.3} & 50.33 & 57.8 & \textbf{60.15} & \textbf{67.3} & 62.85 & 66.7 & 120\,KB \\
\gmmjl & 46.71 & 54.8 & 49.92 & 56.9 & 59.37 & 65.4 & 60.03 & 63.1 & 40\,KB \\
\gmmjlf & 47.28 & 53.5 & 49.68 & 55.2 & 52.23 & 56.7 & \textbf{67.14} & \textbf{69.2} & 40\,KB \\
\kcjl & 45.85 & 51.9 & 48.53 & 55.0 & 53.48 & 58.6 & 49.96 & 53.9 & 50\,KB \\
\chjl & 45.14 & 50.3 & 46.02 & 53.8 & 55.71 & 61.7 & 48.36 & 52.5 & 61\,KB \\
\hil & 46.33 & 54.2 & 45.87 & 53.4 & 57.62 & 63.9 & 59.84 & 63.6 & 120\,KB \\
\hilf & 45.47 & 52.6 & 45.05 & 51.9 & 53.19 & 58.1 & 54.21 & 57.8 & 120\,KB \\
\bottomrule
\end{tabular}}
\end{table}

\begin{table}[t]
\centering
\caption{Geometric descriptor variants for known-vs-novel selection, further benchmarks. Notation as in \cref{tab:selection-primary}.}
\label{tab:selection-extended}
\scriptsize
\setlength{\tabcolsep}{2pt}
\begin{tabular}{l cc cc cc}
\toprule
& \multicolumn{2}{c}{\textbf{DomainNet} ($T{=}4$)} & \multicolumn{2}{c}{\textbf{CUB-200} ($T{=}9$)} & \multicolumn{2}{c}{\textbf{Cars} ($T{=}9$)} \\
\cmidrule(lr){2-3}\cmidrule(lr){4-5}\cmidrule(lr){6-7}
\textbf{Variant} & $A_T$ & $A$ & $A_T$ & $A$ & $A_T$ & $A$ \\
\midrule
\gmm & 43.32 & 44.1 & 79.50 & 82.5 & \textbf{49.60} & \textbf{51.8} \\
\gmmf & 53.36 & 54.4 & \textbf{83.51} & \textbf{86.6} & 48.52 & 50.7 \\
\gmmjlf & 50.27 & 51.3 & 80.20 & 83.2 & 47.85 & 50.0 \\
\hilf & \textbf{56.47} & \textbf{57.6} & 52.27 & 54.2 & 48.83 & 51.0 \\
\hiljlf & 56.11 & 57.2 & 50.92 & 52.8 & 46.44 & 48.5 \\
\chjl & 45.63 & 46.6 & 46.71 & 48.9 & 46.49 & 48.6 \\
\bottomrule
\end{tabular}
\end{table}

\section{Deployment and retained information}
\label{sec:deploy}
Routing requires a frozen backbone pass followed by an adapted pass; likelihood evaluation and grouping by route add overhead.
At each adapted layer, the selected low-rank branches act on the same layer input and their weighted outputs are summed with the frozen layer output; top-3 does not require three complete adapted backbone passes.
Component timings from other configurations do not establish end-to-end latency for this system.

Memory accounting in the main paper describes packed fp32 representations, with only one triangle of each symmetric matrix, and excludes the common frozen backbone, optimizer state, evaluation caches, and temporary sampled features.
It is not measured runtime memory: the current router and ridge evaluator accumulate full fp64 matrices. On CIFAR-100, their scatter-plus-means and Gram-plus-means arrays require approximately $5.33$ and $5.35$\,MB, respectively, before solver workspace or derived prediction weights, compared with $1.49$\,MB each in the packed fp32 accounting.
These are separate accumulators for frozen and adapted features; the ridge moments include a bias coordinate. Together, their base arrays occupy $10.68$\,MB in the evaluator, versus $2.98$\,MB when packed as fp32; neither total measures peak runtime memory.
The carried training head requires $(d+1)C$ fp32 parameters ($0.31$\,MB for $d=768$, $C=100$). Counts and class/task identifiers require $O(C+T)$ integers; the pooled scatter also requires a total sample count.
The ridge prediction weights can be reconstructed from $G$ and $U$; caching them uses another $(d+1)C$ fp32 values. The quoted router storage ratio compares representations under the same packed fp32 convention, not peak memory of the experimental code.
At $T=20$ on CIFAR-100, the replay-free \method\ keeps adapters, class means and the pooled covariance, class statistics, ridge moments, and the training head; the stored-feature run additionally keeps all 50,000 adapted training features.
The auxiliary experiment retains several heads and alignment representations simultaneously; its total experimental memory is larger than any individual head-memory row.

No differential-privacy guarantee is claimed for these experiments.
Fitted distribution parameters and moments can still reveal information about their training data, and the stored-feature run retains training features.
A calibrated privacy guarantee would require a specified neighboring-dataset relation, clipping, sensitivity and composition accounting; a parameter-noise sweep or a weak membership attack alone does not supply one.


\subsection{Inference cost}
\label{sec:latency}
\Cref{alg:infer} runs two backbone passes per input: a frozen pass for routing and an adapted pass for prediction.
\Cref{tab:latency} lists the measured per-sample CPU cost of the components and the end-to-end bound they imply.
The adapted pass performs every operation of the frozen pass and additionally evaluates the selected low-rank branches, so it costs at least one frozen pass.

Timings use the CPU (14 cores) without GPU acceleration, with PyTorch 2.11 and its default thread count, batch size 1. Each run times one CIFAR-100 test input: the per-task GMM router ($T{=}20$, $K{=}100$, NumPy) and the linear head over 200 repetitions, and the backbone over 10 passes of one $224{\times}224$ input after a warm-up pass. Rows report mean $\pm$ standard deviation over four runs. GPU runtimes would be lower because parallelization allows routing to run concurrently, reducing the overhead measured here.

\begin{table}[t]
\centering
\caption{Per-sample CPU inference cost, ViT-B/16, CIFAR-100. Upper block: measured components (mean $\pm$ std). Lower block: derived from the measured rows, not timed end to end. Relative cost is with respect to one frozen backbone pass.}
\label{tab:latency}
\footnotesize
\setlength{\tabcolsep}{3pt}
\begin{tabular}{@{}lrr@{}}
\toprule
\textbf{Component} & \textbf{Time / sample} & \textbf{Relative} \\
\midrule
\multicolumn{3}{@{}l}{\emph{Measured}} \\
Frozen backbone pass & $27.96\pm0.43$\,ms & $1.0\times$ \\
Linear head & $3.34\pm2.72\,\mu$s & ${<}0.001\times$ \\
Per-task GMM routing (ablation router) & $5.83\pm2.85$\,ms & $0.21\times$ \\
\midrule
\multicolumn{3}{@{}l}{\emph{Derived}} \\
Single-pass reference (merged adapters) & $27.96$\,ms & $1.0\times$ \\
Two-pass routed inference & $\sim 47.92$\,ms & $\sim 1.7\times$ \\
\quad with per-task GMM routing & $\sim 53.75$\,ms & $\sim 1.9\times$ \\
\bottomrule
\end{tabular}
\end{table}

{
    \small
    \bibliographystyle{ieeenat_fullname}
    \bibliography{cvpr}
}
\end{document}